\documentclass[journal,10pt,twocolumn,a4]{IEEEtran}

\usepackage{graphicx}
\usepackage{float}
\usepackage{stfloats}
\usepackage{wrapfig}

\usepackage{url} 
\usepackage{amsmath}
\usepackage{amssymb}
\usepackage{cite}
\usepackage{enumitem}
\usepackage{lscape}
\usepackage{rotating}
\usepackage{tabularx}
\usepackage{subfiles}
\usepackage{epstopdf}

\usepackage[caption=false, font=footnotesize]{subfig}

\usepackage{multirow}
\usepackage{algorithmic}
\usepackage{hyperref}
\usepackage{acronym}
\usepackage{afterpage}
\usepackage{pdfpages}
\usepackage{blindtext}
\usepackage{array}
\usepackage{xcolor}
\usepackage{gensymb}

\usepackage{algorithm,algorithmic}

\definecolor{darkblue}{rgb}{0, 0, 153}

\DeclareMathOperator*{\argmax}{\arg\!\max}
\DeclareGraphicsExtensions{.pdf,.png,.jpg}

\begin{document}

\setlength{\abovedisplayskip}{7pt}
\setlength{\belowdisplayskip}{7pt}
\raggedbottom

\title{Detecting Dynamic Community Structure in Functional Brain Networks Across Individuals: A Multilayer Approach}

\author{Chee-Ming~Ting,
        S.~Balqis~Samdin,
        Meini~Tang,
        and~Hernando~Ombao
\thanks{C.-M. Ting is with the School of Information Technology, Monash University Malaysia, 47500 Subang Jaya, Malaysia, and also the Biostatistics Group, King Abdullah University of Science and Technology, Thuwal 23955, Saudi Arabia (e-mail: ting.cheeming@monash.edu).}
\thanks{S. B. Samdin is with the School of Electrical and Computer Engineering, Xiamen University Malaysia, 43900 Sepang, Malaysia, and also the Biostatistics Group, King Abdullah University of Science \& Technology, Thuwal 23955, Saudi Arabia.}
\thanks{M. Tang \& H. Ombao are with the Biostatistics Group, King Abdullah University of Science \& Technology, Thuwal 23955, Saudi Arabia.
}
}

\markboth{}%
{Shell \MakeLowercase{\textit{et al.}}: Bare Demo of IEEEtran.cls for Journals}

\maketitle

\begin{abstract}
\textit{Objective}: We present a unified statistical framework for characterizing community structure of brain functional networks that captures variation across individuals and evolution over time. Existing methods for community detection focus only on single-subject analysis of dynamic networks; while recent extensions to multiple-subjects analysis are limited to static networks.
\textit{Method}: To overcome these limitations, we propose a multi-subject, Markov-switching stochastic block model (MSS-SBM) to identify state-related changes in brain community organization over a group of individuals. We first formulate a multilayer extension of SBM to describe the time-dependent, multi-subject brain networks. We develop a novel procedure for fitting the multilayer SBM that builds on multislice modularity maximization which can uncover a common community partition of all layers (subjects) simultaneously. By augmenting with a dynamic Markov switching process, our proposed method is able to capture a set of distinct, recurring temporal states with respect to inter-community interactions over subjects and the change points between them. \textit{Results}: Simulation shows accurate community recovery and tracking of dynamic community regimes over multilayer networks by the MSS-SBM. Application to task fMRI reveals meaningful non-assortative brain community motifs, e.g., core-periphery structure at the group level, that are associated with language comprehension and motor functions suggesting their putative role in complex information integration. Our approach detected dynamic reconfiguration of modular connectivity elicited by varying task demands and identified unique profiles of intra and inter-community connectivity across different task conditions. \textit{Conclusion}: The proposed multilayer network representation provides a principled way of detecting synchronous, dynamic modularity in brain networks across subjects.
\end{abstract}


\IEEEpeerreviewmaketitle

\vspace{-0.15in}

\section{Introduction}

\IEEEPARstart{F}{unctional} architecture of the brain can be characterized as a network of interconnected regions. Study of brain networks has offered new insights on human behavior and neurodegenerative diseases \cite{Bullmore2009}. Early studies using functional magnetic resonance imaging (fMRI) assume a static functional connectivity (FC) pattern over time. Recent evidence suggests temporal dynamics of FC patterns over multiple time scales during task performance and rest \cite{Hutchison2013}. Dynamic FC has also been studied to examine the normal and pathological brain connectivity patterns \cite{Rashid2014}. Despite dynamic fluctuations over time, FC tends to be temporally clustered into a finite number of putative connectivity states, i.e., distinct connectivity patterns that transiently recur over the course of experiment \cite{Baker2014, Allen2012}. Most studies of dynamic connectivity states focused on transition between whole-brain connectivity profiles only in terms of connectivity edges. However, switching in the topological properties of brain functional networks such as the modular or community structure has received less attention. Our goal is to develop a novel approach to quantifying dynamic FC, specifically the state-driven changes in community organization of brain networks, while also taking into account variation across individuals.

Evidence from neuroimaging studies suggests complex community structure of both structural and functional brain networks \cite{Sporns2016}, where brain network can be decomposed into clusters of densely inter-connected nodes (called modules or communities) that are relatively sparsely connected with nodes in other modules. These topological modules often correspond to groups of anatomically neighboring and/or functionally-related brain regions that are engaged in specialized information processing. Many data-driven community detection methods have been applied to identify latent community structure in brain networks. The most widely-used approach is the modularity maximization which partitions network's nodes into non-overlapping communities that are more internally dense than would be expected by chance, by maximizing an objective function of modularity \cite{Newman2004}. There are many computationally-efficient heuristics that search for the approximate optimal modularity \cite{Clauset2004}. Among them is the popular Louvain algorithm which is the fastest community detection methods in practice \cite{Blondel2008} but is only suited for analysis of single-layer networks, e.g., for individual subjects.

We consider a statistically-principled approach using the stochastic block model (SBM), a generative model for networks with community structure \cite{Nowicki2001}. The SBM partitions a network into `blocks' or communities of nodes such that the probability of forming a connectivity edge between a pair of nodes depends only on which communities these nodes belong. One advantage of SBM is that it offers a richer class of community structures beyond the traditional assortative community with internally dense and externally sparse connections (i.e., the probability of an edge between nodes is higher within a community than between communities). These non-assortative structures include the core-periphery, disassortative and mixed motifs \cite{Betzel2018}. The maximizers of Newman-Girvan modularity \cite{Newman2004} have been proven as asymptotically consistent estimators of block partitions under the SBM \cite{Bickel2009}, and recently extended to degree-corrected SBM \cite{Chen2018}.

Despite that community detection has become important for brain network analysis, there has not been much progress in (1) quantifying dynamic changes in community structure over time, and (2) detecting and mapping communities across subjects. Modularity in functional networks can exhibit changes across time, e.g., over the course of task performance and learning \cite{Bassett2011,Pedersen2018}. Most studies using SBMs for brain networks focused mainly on the static descriptions of functional brain modules \cite{Pavlovic2014,Betzel2018}. Extensions of SBM for static networks to dynamic settings have been introduced recently to detect temporal evolution of communities in social networks \cite{Xu2014,Matias2017}. To our knowledge, application of dynamic SBMs to time-varying brain networks is still very limited. Detecting brain community structure across subjects was traditionally performed based on individual subjects or group-averaged networks. This approach however suffers from inconsistent mapping of community labels across subjects and relies on some ad-hoc template-matching techniques to register the subject-specific communities to a common template \cite{Gordon2017}.

A recent solution to these problems is the multilayer network representation via aggregating multiple instances of a single network (layers) and then identifying communities across layers by maximizing a multilayer modularity function \cite{Mucha2010}. A few studies have applied this approach to time-varying brain networks to track changes in community assignments of nodes across time \cite{Bassett2011,Pedersen2018}, where each layer represents a snapshot of functional network at a particular time window with inter-layered couplings to connect nodes of networks between adjacent time points. It was recently modified to characterize modularity in brain networks across subjects \cite{Betzel2019}. By applying modularity maximization to a multilayer stack of individual subjects' connectivity matrices, it can find communities in all layers (i.e. subjects) simultaneously. One advantage is that it preserves community labels that are consistent across different subjects, thus allowing straightforward inter-subject mapping of community assignments.

In this paper, we extend the SBM to dynamic, multi-subject networks and adopt the multilayer modularity for detecting communities. Specifically, we develop a novel framework based on multi-subject, Markov-switching SBM (MSS-SBM) to identify dynamic changes in modular organization of brain networks across subjects. We first formulate a multilayer SBM to characterize community structure in multi-subject, time-varying brain functional networks. We leverage on the multilayer modularity maximization to find shared community partition across subjects. Secondly, we aim to detect state-based changes in the network modular organization, i.e., distinct patterns of inter-modular connectivity that repetitively occur over time and across subjects, driven by some latent brain states in response to changes in task conditions or stimuli over course of experiments. By combining the multilayer SBM and a hidden Markov model (HMM) to describe the evolution of the underlying states, the proposed MSS-SBM is able to estimate simultaneously the change-points of time-evolving modularity states and the block structure in each state, i.e., intra- and inter-modular connections. It is flexible to capture a variety of dynamics, e.g., a shift from a connectivity state which is highly modular to a state which is less modular and more integrated throughout the network. Moreover, our model does not require for the timing of the switching between states to be known a priori. In contrast to a similar setup in \cite{Robinson2015} that uses hidden-Markov SBM on the observed time-varying graphs directly, our approach has the advantage of identifying distinct temporal states in dynamic community structure based on lower-dimensional, time-evolving inter-modular connectivity matrices. Moreover, \cite{Robinson2015} only analyzed group-averaged dynamic functional networks and neglected variation across subjects. A multi-subject SBM based on mixture modeling and variational Bayesian estimation was recently proposed by \cite{Pavlovic2019}, which however did not address the dynamic nature of the modular organization. Our earlier work \cite{Balqis2019} proposed a Markov-switching SBM which revealed alternating modular connectivity in fMRI functional networks during language processing, but it uses spectral clustering for community detection and is limited to single-subject analysis. We apply the proposed MSS-SBM to task fMRI data in Human Connectome Project (HCP) to study rapid switching of brain network modularity evoked by repetitive tasks.

The main contributions of this work are as follows:
\begin{itemize}
\item[1)] We propose a novel framework based on MSS-SBM to characterize state-based dynamic community structure of brain functional networks across subjects.
\item[2)] Our method combines a multi-subject, time-varying SBM with an HMM to identify distinct repeating states in the time-varying inter-community connectivity without a priori knowledge about the timing of the structural switching between these states of network modularity.
\item[3)] To the best of our knowledge, our proposed approach is the first that leverages on the multilayer modularity maximization to detect community structure of brain networks in multiple subjects simultaneously under the proposed MSS-SBM. Given the common community partition with consistent mapping of nodes' community assignments across subjects, it allows us to identify a set of group-level connectivity states.
\end{itemize}

\vspace{-0.1in}
\section{Modeling Multi-Subject Dynamic Community Structure in Brain Networks}
We first describe a novel multilayer SBM for modeling community structure in multi-subject, time-varying brain functional networks. To identify state-related changes in the time-evolving community structure, we further develop a MSS-SBM that combines the multilayer SBM with an HMM to describe the switching between distinct states of modular connectivity patterns over time and across subjects. The notations of the proposed model are given in Table.~\ref{Table:Notation}.

\begin{table*}[!t]
\renewcommand{\arraystretch}{1.2}
\caption{Overview of notations of the proposed MSS-SBM}
\vspace{-0.2 cm}
\label{Table:Notation}
\centering
\resizebox{0.8\textwidth}{!}{
\begin{tabular}{ll}
  \hline \hline
Notation & Description \\
\hline
$T$, $R$, $N$ & Number of time points, number of subjects, number of nodes\\
$G^{r,t} \equiv \lbrace V, E^{r,t}\rbrace$ & Brain networks at time $t$ for $r$th subject\\
$V \equiv \lbrace V_1,\ldots, V_N \rbrace$, $E^{r,t} \equiv \lbrace e^{r,t}_{ij}\rbrace$ & Set of nodes, set of edges between nodes in $G^{r,t}$\\
$\mathbf{W}^{r,t} = [w_{ij}^{r,t}] \in {\lbrace 0,1 \rbrace}^{N \times N}$ & Adjacency matrix for $G^{r,t}$ ($w_{ij}^{r,t} = 1$ if node $i$ links to node $j$, $0$ otherwise)\\
\hline
$K$ & Number of communities (or modules)\\
$N_k$ & Number of nodes in $k$th community\\
$g_i \in \lbrace 1,\dots, K\rbrace$ & Community label of $i$th node\\
$\phi^{r,t}_{i j} \in [0,1]$ & Probability of connections between nodes $i$ and $j$  at time $t$ for $r$th subject\\
$\theta^{r,t}_{k l} \in [0,1]$ & Probability of connections between nodes in communities $k$ and $l$\\
$\mathbf{g} = (g_1, \ldots, g_N)$ & Community membership vector\\
$\boldsymbol{\Omega} = [\omega_{ik}] \in {\lbrace 0,1 \rbrace}^{N \times K}$ & Community membership matrix ($\omega_{i,g_i} = 1$ and $0$ elsewhere)\\
$\boldsymbol{\Phi}^{r,t} = [\phi^{r,t}_{i j}] \in [0,1]^{N \times N}$ & Node-wise connection probability matrix \\
$\boldsymbol{\Theta}^{r,t} = [\theta^{r,t}_{k l}] \in [0,1]^{K \times K}$ & Module-wise connection probability matrix \\
$\boldsymbol{\beta}^{r,t} \in \mathbb{R}^{K^2}$ & Vectorized logit transform of $\boldsymbol{\Theta}^{r,t}$\\
\hline
$S$ & Number of states\\
$s_{r,t} \in \lbrace 1,\dots, S\rbrace$ & State indicator at time $t$ for $r$th subject\\
$\pi_{\ell m} \in [0,1]$ & Transition probability from state $\ell$ to state $m$\\
$\mathbf{\Pi} =[\pi_{\ell m}] \in [0,1]^{S \times S}$ & Transition probability matrix\\
$\boldsymbol{\mu}^{[m]}_{\Theta} \in \mathbb{R}^{K^2}$, $\boldsymbol{\Sigma}^{[m]}_{\Theta} \in \mathbb{R}^{K^2 \times K^2}$ & Mean vector and covariance matrix of modular connectivity $\boldsymbol{\Theta}$ for $m$th state\\
\hline \hline
\end{tabular}
}
\vspace{-0.05in}
\end{table*}

\vspace{-0.1in}
\subsection{Multilayer SBM} \label{MethodDSBM}

We consider a collection of undirected graphs of multi-subject, time-varying functional brain networks $\mathcal{G} = \lbrace G^{r,t}, t=1, \ldots, T,  r=1, \ldots, R \rbrace$ that share a set of nodes $V \equiv \lbrace V_1,\ldots, V_N \rbrace$ (voxels or regions of interest (ROIs)) over $T$ time points for a group of $R$ subjects. We can view $\mathcal{G}$ as a doubly-indexed multilayer network where each (${r,t}$)th layer $G^{r,t} \equiv \lbrace V, E^{r,t}\rbrace$ represents a snapshot of a network observed at time step $t$ for the $r$th subject, with a set of (possibly time-changing) connectivity edges between $N$ individual nodes denoted by $E^{r,t} \equiv \lbrace e^{r,t}_{ij}, 1\leq i,j \leq N \rbrace$. We assume the number of brain nodes $N=|V|$ to be fixed over time and subjects. We define the corresponding adjacency matrix representations of the multi-subject, time-dependent networks in $\mathcal{G}$ by $\mathbf{W} = \lbrace \mathbf{W}^{r,t}, t=1, \ldots, T,  r=1, \ldots, R \rbrace$ where $\mathbf{W}^{r,t} = [w_{ij}^{r,t}]$ is a $N\times N$ symmetric matrix at time $t$ for subject $r$ with $w_{ij}^{r,t} = 1$ if there exists a connecting edge between the nodes $i$ and $j$, $e^{r,t}_{ij} \in E^{r,t}$ and $w_{ij}^{r,t} = 0$ otherwise. We assume there is no self-edge, i.e., $w^{r,t}_{ii} = 0$. The time-varying adjacency matrices for each subject can be estimated by thresholding the dynamic FC matrices (e.g., sliding-window correlation matrices).

Under multilayer SBM, functional networks in $\mathcal{G}$ are assumed to be generated from a set of SBMs, where $\mathbf{W}^{r,t}$ of individual layers follows a regular single-layer SBM which partitions the $N$ network nodes into $K$ blocks or communities (clusters of anatomically or functionally-related brain regions). Let $\mathbf{g} = (g_1, \ldots, g_N)$ be $N \times 1$ community membership vector, where $g_i \in \lbrace 1,\dots, K\rbrace$ indicates the community membership label of node $V_i$ and $g_i=k$ if node $V_i$ belongs to community $k$. We also denote $\Gamma_k = \Gamma_k(\mathbf{g}) = \lbrace V_i : g_i=k \rbrace$ and $N_k = |\Gamma_k|$ to be the set of nodes and number of nodes within community $k$ for $k = 1, \ldots, K$. We can rewrite in a $N \times K$ membership matrix $\boldsymbol{\Omega} = [\omega_{ik}]$ such that $i$th row of $\boldsymbol{\Omega}$ is 1 in the $g_i$th column, $\omega_{i,g_i} = 1$ and $0$ elsewhere. Each node belongs only to one community (i.e., the communities or blocks are disjoint) such that $\sum_{k=1}^K \omega_{ik} = 1$. We also define a $K \times K$ symmetric modular connection probability matrix $\boldsymbol{\Theta}^{r,t} = [\theta^{r,t}_{k l}]$, where $\theta^{r,t}_{k l} \in [0,1]$ is the probability of edges existing between any node in community $k$ and any node in community $l$ at time $t$ for subject $r$. The diagonal elements $\theta^{r,t}_{kk}$ and the off-diagonals $\theta^{r,t}_{kl}, k\neq l$ capture the within-module and between-module connectivity, respectively. Conditioned on the community assignments of nodes $g_i$ and $g_j$, edges within and across network layers (${r,t}$) are formed independently following a Bernoulli distribution
\begin{equation}
{w}_{ij}^{r,t} \sim \textit{Bernoulli} (\phi_{ij}^{r,t}) \label{Bernoulli-edge}
\end{equation}
where $\phi_{ij}^{r,t} = \theta^{r,t}_{g_i g_j}$. Note that the probability of a connection $\phi_{ij}^{r,t} = Pr({w}_{ij}^{r,t} = 1)$ between nodes $i$ and $j$ depends only on the community blocks to which they belong. Then, the node-wise connectivity matrix is defined by $\boldsymbol{\Phi}^{r,t} = [\phi^{r,t}_{i j}] = \boldsymbol{\Omega}\boldsymbol{\Theta}^{r,t}\boldsymbol{\Omega}^T$. Model (\ref{Bernoulli-edge}) assigns a separate connection probability for each subject and each time point. The set of parameters of the multilayer SBM is denoted by $\lbrace \mathbf{g}, \boldsymbol{\Theta} \rbrace$ with $\boldsymbol{\Theta} = \lbrace \boldsymbol{\Theta}^{r,t}; t=1, \ldots, T,  r=1, \ldots, R\rbrace$. In our setting, the network community partition as represented by $\mathbf{g}$ is assumed to be common to all subjects and constant over time, but the modular connectivity matrix $\boldsymbol{\Theta}^{r,t}$ is allowed to evolve across time and to vary across subjects. We consider estimation of multilayer SBM with $K$ blocks in the \textit{a posteriori} setting where both the community membership labels $\mathbf{g}$ and the connectivity matrices $\boldsymbol{\Theta}$ are both unknown and to be estimated.

\vspace{-0.1in}
\subsection{Multi-Subject Markov-Switching SBM}

In contrast to recent studies of dynamic connectivity states in the whole-brain connectivity edges \cite{Allen2012,Bandettini2015,Samdin2017,Ting2018}, our goal in this paper is to identify distinct states in the time-evolving modular organization of networks and the temporal locations of transitions between states. We develop a regime-switching SBM to characterize changes the inter-community connectivity driven by a set of recurring latent states over time and subjects. In particular, let $\boldsymbol{\beta}^{r,t} = \text{vec}(g(\boldsymbol{\Theta}^{r,t}))$ be $K^2$-dimensional vectorized version of $g(\boldsymbol{\Theta}^{r,t})$ and $g(\boldsymbol{\Theta}^{r,t}) = \text{logit}(\boldsymbol{\Theta}^{r,t})$ whose elements are logit of $\theta^{r,t}_{k l}$, $\text{logit}(\theta^{r,t}_{k l}) = \log(\theta^{r,t}_{k l}) - \log(1-\theta^{r,t}_{k l})$. We assume the logit transform of time-varying modular connection probabilities $\boldsymbol{\Theta}^{r,t}$ in (\ref{Bernoulli-edge}) to follow an HMM
\begin{align}
s_{r,t}|s_{r,t-1} = \ell & \sim Multi(\pi_{\ell 1}, \ldots, \pi_{\ell S}) \label{eq:HMM-state} \\
\boldsymbol{\beta}^{r,t}|s_{r,t} = m & \sim N(\boldsymbol{\mu}^{[m]}_{\Theta}, \boldsymbol{\Sigma}^{[m]}_{\Theta}) \label{eq:HMM-obs} \\
\mathbf{W}^{r,t}|\mathbf{g},\boldsymbol{\Theta}^{r,t} & \sim Bernoulli (\boldsymbol{\Omega}\boldsymbol{\Theta}^{r,t}\boldsymbol{\Omega}^T) \label{eq:Bernuolli-matrix} 
\end{align}
where $s_{r,t} \in \lbrace 1,\dots, S\rbrace$ for $t = 1,\ldots, T$ is a sequence of state variables which vary over time for $r$th subject, $S$ is the number of states. The variation in the modular connectivity structure over time and subjects is determined by the latent state indicator $s_{r,t}$ which follows a Markov process with $S \times S$ transition matrix $\mathbf{\Pi} =[\pi_{\ell m}]_{1 \leq \ell, m \leq S}$, where $\pi_{\ell m} = Pr(s_{r,t}=m \vert s_{r,t-1} = \ell)$ is the probability of transition from state $\ell$ at time $t-1$ to state $m$ at time $t$. The parameters $\boldsymbol{\mu}^{[m]}_{\Theta}$ and $\boldsymbol{\Sigma}^{[m]}_{\Theta}$ capture respectively the mean and variations of inter-modular connection probabilities in each state $m = 1,\dots, S$.
In analyzing group-wise time-varying networks averaged over subjects, \cite{Robinson2015} fitted the hidden Markov SBM directly on the high-dimensional $N \times N$ node-wise connectivity matrices $\mathbf{W}^{t}$, specifying the evolution of connectivity parameters $\boldsymbol{\Theta}^{[s_t]}$ as a piecewise constant function of $s_t$. In contrast, the advantage of our approach is that it utilizes the HMM for $K \times K$ modular connectivity matrices $\boldsymbol{\Theta}^{r,t}$, which involves a smaller number of parameters in the state estimation and thus improving computational and statistical efficiency. Moreover, it allows clustering of the time-evolving community structure into states that maybe associated with different tasks and conditions over the time course of experiment. Given the model (\ref{eq:HMM-state})-(\ref{eq:Bernuolli-matrix}), the aims are to estimate the state sequence $s_{r,t}$ which indicates which regime to be most likely active at each time point and for each subject, and the state-specific modular connectivity parameters $\lbrace \boldsymbol{\mu}^{[m]}_{\Theta}, \boldsymbol{\Sigma}^{[m]}_{\Theta}, m= 1,\dots, S \rbrace$.

\begin{figure*}[!t]
\centering
\includegraphics[width=0.95\linewidth,keepaspectratio]{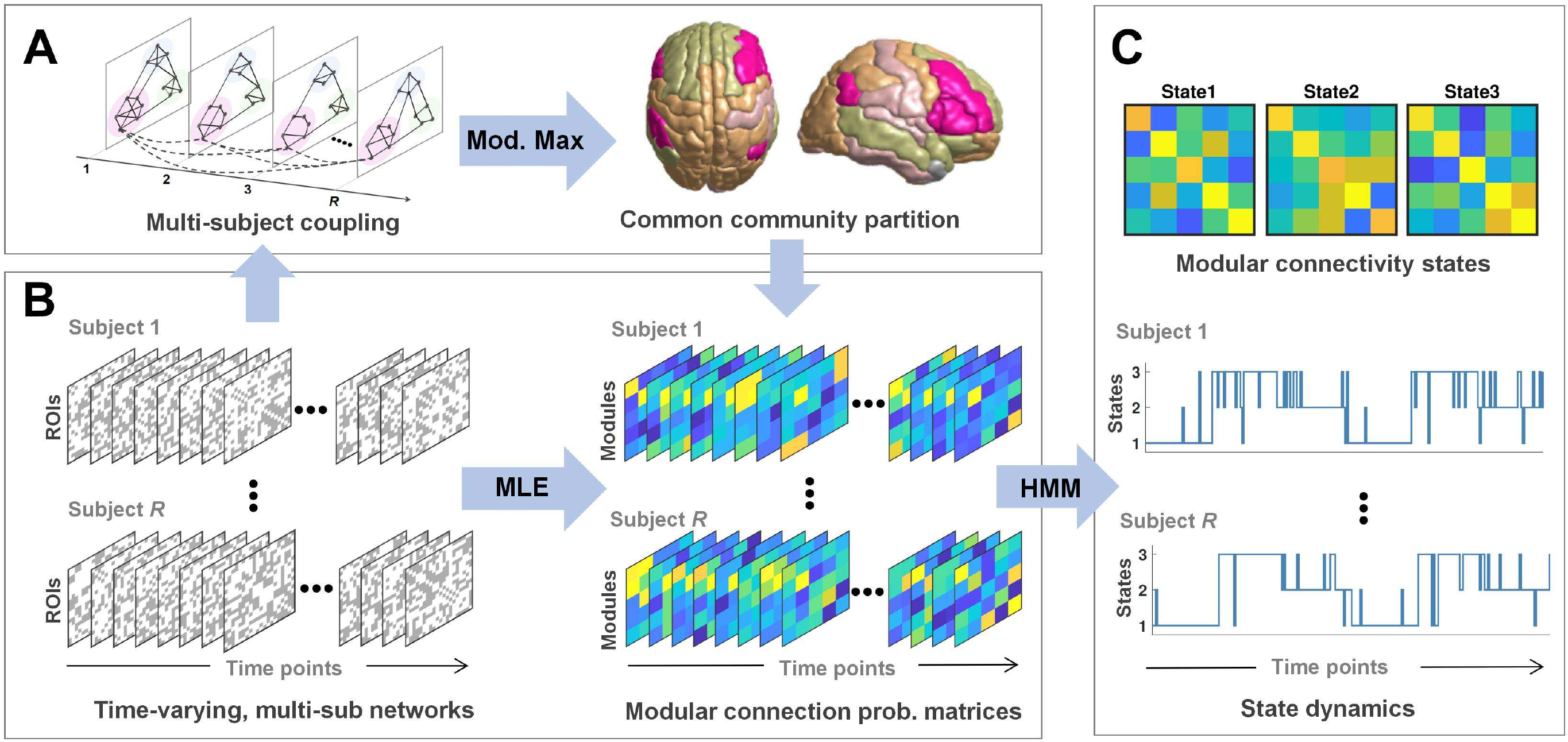}
\vspace{-0.05in}
	 \caption{Overview of the proposed MSS-SBM framework for detecting state-based dynamic community structure in multi-subject brain functional networks. The approach consists of three components: (A) Community detection via multilayer modularity maximization. (B) Estimation of time-resolved inter-community connectivity. (C) Identification of dynamic modular connectivity states via hidden Markov modeling.}  \label{block-diag}
	\vspace{-0.1in}
\end{figure*}

\section{Estimation}

We develop a unified framework for efficient estimation of the proposed MSS-SBM to identify state-based dynamic community structure in multiple subjects. A schematic overview is shown in Fig.~\ref{block-diag}. The estimation consists of two steps:

\textit{Stage 1}: Fit the multilayer SBM to the multi-subject, time-varying adjacency matrices in $\mathbf{W}$. We first estimate the common block structure $\mathbf{g}$, by applying the modularity maximization algorithm to a group-level multilayer network object comprising single networks of individual subjects to uncover the shared nodes community memberships $\mathbf{g}$ over all subjects simultaneously. Given the estimated community partition, we then estimate by maximum likelihood (ML) method the inter-modular connection probabilities $\lbrace \boldsymbol{\Theta}^{r,t} \rbrace$ for each subject and each time point based on $\lbrace \mathbf{W}^{r,t} \rbrace$. 

\textit{Stage 2:} Fit the HMM on $\lbrace \boldsymbol{\Theta}^{r,t} \rbrace$ to identify dynamic community states. This step produces estimates of change-points between states across time and subjects via the mostly likely state sequence $s_{r,t}$ and the state-specific modular connectivity parameters $\{\boldsymbol{\mu}^{[m]}_{\Theta},\boldsymbol{\Sigma}^{[m]}_{\Theta}\}$. The proposed estimation procedure is summarized in Algorithm 1.

\begin{algorithm}[!t]
\caption{Estimation Algorithm for MSS-SBM}
\begin{algorithmic}[1]
\renewcommand{\algorithmicrequire}{\textbf{Input:}}
\renewcommand{\algorithmicensure}{\textbf{Output:}}
\vspace{0.06in}
\REQUIRE Multi-subject, time-varying and time-averaged adjacency matrices $\lbrace \mathbf{W}^{r,t}, t=1, \ldots, T,  r=1, \ldots, R \rbrace$ and $\lbrace \overline{\mathbf{W}}^1, \ldots, \overline{\mathbf{W}}^R\rbrace$.

\vspace{0.05in}
\renewcommand{\algorithmicrequire}{\textbf{Parameters:}}
\REQUIRE Structural resolution $\gamma_r = \gamma = 1$, inter-layer coupling $C_{jrs} = C = 1$.

\vspace{0.06in}
\renewcommand{\algorithmicrequire}{\textit{Step 1: Community Detection}}
\REQUIRE \vspace{0.04in}
\STATE Find membership vector $\widehat{\mathbf{g}}$ by applying generalized Louvain algorithm \cite{Jeub2017} to maximize multilayer modularity (\ref{eq:MSQ-function})

\vspace{0.05in}
\renewcommand{\algorithmicrequire}{\textit{Step 2: Modular Connectivity Estimation}} 
\REQUIRE \vspace{0.04in}
\STATE Set $\widehat{N}_k = |\{V_i:\widehat{g}_i=k\}|$ for $k=1, \ldots, K$ \vspace{0.03in}
\STATE Set $n_{kl}=\widehat{N}_k \widehat{N}_l$ for $k \neq l$ and $n_{kk}=\widehat{N}_k(\widehat{N}_k-1)$ \vspace{0.03in}
\FOR {$r = 1:R$}
	\FOR {$t = 1:T$}
				\STATE Compute modular connectivity $\widehat{\boldsymbol{\Theta}}^{r,t} = [\hat{\theta}_{k,l}^{r,t}]$ via (\ref{eq:mle})
	\ENDFOR
\ENDFOR

\vspace{0.05in}
\renewcommand{\algorithmicrequire}{\textit{Step 3: Dynamic State Identification}}
\REQUIRE \vspace{0.04in}
\STATE Concatenate $(\widehat{\boldsymbol{\beta}}^{1,1}, \ldots, \widehat{\boldsymbol{\beta}}^{R,T})$ with $\widehat{\boldsymbol{\beta}}^{r,t} = \text{vec}(g(\widehat{\boldsymbol{\Theta}}^{r,t}))$  \vspace{0.03in}
\STATE Compute state connectivity $\{\widehat{\boldsymbol{\mu}}^{[m]}_{\Theta}, \widehat{\boldsymbol{\Sigma}}^{[m]}_{\Theta}\}$ and transition probability $\widehat{\boldsymbol{\Pi}}$ by fitting an HMM to $(\widehat{\boldsymbol{\beta}}^{1,1}, \ldots, \widehat{\boldsymbol{\beta}}^{R,T})$ using the EM algorithm \vspace{0.03in}
\STATE Generate state sequence $\widehat{s}_{1,1},\ldots, \widehat{s}_{R,T}$ by solving (\ref{eq:VB}) using the Viterbi algorithm \vspace{0.03in}

\vspace{0.06in}
\ENSURE $\lbrace \widehat{\mathbf{g}}, \widehat{\boldsymbol{\Theta}}\rbrace$, $\{\widehat{\boldsymbol{\Pi}}, \widehat{\boldsymbol{\mu}}^{[m]}_{\Theta}, \widehat{\boldsymbol{\Sigma}}^{[m]}_{\Theta}\}$ and $\{\widehat{s}_{r,t}\}$.
\end{algorithmic}
\end{algorithm}

\vspace{-0.3in}
\subsection{Community Detection} \label{MethodComDetect}

To detect communities in multi-subject brain functional networks, we develop a method inspired by the modularity maximization ($\text{Q}_{\max}$) approach for estimating the multilayer SBM. The $\text{Q}_{\max}$ algorithm provides an estimate of the number of communities $K$ which is then used for fitting the SBM. We use the single-layer generalized Louvain algorithm to estimate the community membership of nodes $\mathbf{g}$ at the individual subject level, and develop an extension of the $\text{Q}_{\max}$ to multilayer networks for group-level analysis.

\subsubsection{Single-Layer Modularity Maximization} \label{LouvainMethod}
Let $\overline{\mathbf{W}}^{r} = [\bar{w}_{ij}^{r}]$ be the adjacency matrix for subject $r = 1,\ldots, R$ which is obtained by thresholding the time-averaged correlation matrix. For single-subject community detection, we aim to find the optimal community membership vector $\mathbf{g}_r = \lbrace g_{1r}, \dots, g_{Nr} \rbrace$ for each subject $r$ independently. This can be accomplished by maximizing the modularity quality function \cite{Newman2006} of single-layer network, defined for each subject as
\vspace{-0.02in}
\begin{equation}
Q(\mathbf{g}_r) = \sum_{i,j} \left(\bar{w}_{ij}^{r} - p_{ijr}\right) \delta(g_{ir},g_{jr}) \label{eq:Q-function}
\end{equation}
where $p_{ijr} = \frac{\kappa_{ir} \kappa_{jr}}{2 L_r}$ denotes the expected weight of the edges connecting nodes $i$ and $j$ under the Newman-Girvan null model, $\kappa_{ir} = \sum_j \bar{w}^r_{ij} $ is the degree of node $i$, $L_r$ is total number of edges in the network of subject $r$, and $\delta(g_{ir},g_{jr}) = 1$ if nodes $i$ and $j$ belong to the same community, and 0 otherwise. Then, the modularity maximization estimator of the community partition is defined by $\hat{\mathbf{g}}_r = \argmax_{\mathbf{g}_r} Q(\mathbf{g}_r)$. The partition that gives the greatest value of $Q$ is considered as a good estimate of a network's community structure. We will drop the subject index $r$ for notational brevity.

To solve the single-subject or single-layer $\text{Q}_{\max}$, we employ the Louvain algorithm which is simple and computationally-efficient. This community detection algorithm aims to find communities in a network assuming that connectivity between nodes within communities is stronger than connectivity between nodes across communities. It is a two-step iterative algorithm. As initialization each node in the network is its own community. In first step, each node will be assigned to a community of neighboring nodes if the resulting network modularity $Q$ is maximized. The gain in modularity $\Delta Q$ by moving a node $i$ into community $k$ is given by \cite{Blondel2008}
\vspace{-0.1in}
\begin{multline}
\Delta Q = \left[ \frac{\Sigma_{in} + \kappa_{i,in}}{2L} - \left( \frac{\Sigma_{tot} + \kappa_{i}}{2L} \right)^2 \right] \\ - \left[ \frac{\Sigma_{in}}{2L} - \left( \frac{\Sigma_{tot}}{2L} \right)^2 - \left( \frac{\kappa_{i}}{2L} \right)^2 \right]  \label{ModularGain}
\end{multline}
where $\Sigma_{in} = \sum_{ij \in \Gamma_k} \bar{w}_{ij}$ is the number of edges within community $k$, $\Sigma_{tot} = \sum_{i \in \Gamma_k} \kappa_i$ is the total number of edges incident to nodes of community $k$, $\kappa_i$ is the degree of node $i$ and $\kappa_{i,in}=\sum_{j \in \Gamma_k} \bar{w}_{ij}$ is number of edges from node $i$ to other nodes in the community $k$. The second step involves constructing a network with the new community structure detected in first step. The two steps are repeated iteratively until convergence of the network modularity. The algorithm may produce different number of communities and community partitions $\mathbf{g}_r$ across subjects.

\subsubsection{Multilayer Modularity Maximization}
We apply the multilayer modularity approach \cite{Mucha2010} for community detection in the multi-subject networks to find a group-level community partition. With multilayer modularity optimization, one can study the dynamic network organization over a set of temporally-linked time-dependent networks \cite{Bassett2013, Khambhati2018, Sporns2016, Betzel2018}. This method will identify the nodes community memberships of the functional connectivity networks for all subjects simultaneously. The advantage of this approach is that it can determine consistent community labels for all nodes in multi-subject networks. In contrast, conventional single-layer community detection methods such as spectral clustering suffer from problem of arbitrary community label switching and hence inconsistent mapping of nodes assignments across different subjects \cite{Rohe2011, Jing2015, Matias2017}.

Let $\overline{\mathbf{W}} = \lbrace \overline{\mathbf{W}}^1, \ldots, \overline{\mathbf{W}}^R\rbrace$ be the set of $R$ subject-specific adjacency matrices observed for a multilayer network where each layer represents a static functional brain network of a particular subject. The multilayer modularity across all pairs of subjects $r$ and $s$ is written as
\begin{equation}
Q_{MS} = \frac{1}{2\mu} \sum_{ijrs} [(\bar{w}_{ij}^r - \gamma_r p_{ijr})\delta(g_{ir},g_{jr}) + \delta(i,j)C_{jrs}] \delta(g_{ir},g_{js}) \label{eq:MSQ-function}
\vspace{-0.02in}
\end{equation}
where $g_{ir}$ is the community assignment of node $i$ in layer $r$, $\delta(g_{ir},g_{js})= 1$ indicates that community assignments $g_{ir}$ and $g_{js}$ are identical, and $p_{ijr} = \frac{\kappa_{ir} \kappa_{jr}}{2 L_r}$ is the expected weight of edges within layer $r$. The total number of edges in the adjacency tensor $\overline{\mathbf{W}}$ is $\mu = 0.5 \sum_{js} (\kappa_{js} +c_{js})$ where $c_{js} = \sum_r C_{jrs}$ is the interlayer strength of node $j$ in layer $s$. This modularity optimization depends on intra-layer structural resolution $\gamma$ and interlayer coupling parameter between layers $r$ and $s$ of the same nodes $j$, $C_{jrs}$. By adding interlayer connections of weight $C$, optimization of (\ref{eq:MSQ-function}) yields community labels that are preserved across subjects. We used all-to-all interlayer coupling since the network layers across subjects do not reflect specific order \cite{Mucha2010}. Larger values of $\gamma$ result in many small communities while large values of $C$ produce communities that are common across subjects, with estimated membership vectors $\hat{\mathbf{g}}_r = \hat{\mathbf{g}}$ for all $r=1,\ldots, R$.

\vspace{-0.05in}
\subsection{Estimation of Modular Connectivity Parameters}
Given the estimated community partition, we can estimate the subject-specific, time-dependent modular connectivity parameters $\lbrace \mathbf{\boldsymbol{\Theta}}^{r,t} \rbrace$ via maximum likelihood.
Under the independent formation of edges according to (\ref{Bernoulli-edge}), for any arbitrary community assignment $\mathbf{g}$, the log-likelihood of the set of $R \times T$ adjacency matrices $\mathbf{W}$ under the multilayer SBM is
\begin{flalign}
f({\bf W};\boldsymbol{\Theta},{\bf g}) & = \log \left( \prod_{r=1}^R \prod_{t=1}^T \prod_{i<j} \left(\phi_{ij}^{r,t}\right)^{w_{ij}^{r,t}} \left(1-\phi_{ij}^{r,t}\right)^{1-w_{ij}^{r,t}} \right) \notag \\
 & = \sum_{r=1}^R \sum_{t=1}^T \sum_{i<j} \left\{ w_{ij}^{r,t} \log \theta_{g_i,g_j}^{r,t} \right. \notag \\
 & \hspace{0.7in} \left. + (1-w_{ij}^{r,t}) \log (1-\theta_{g_i,g_j}^{r,t}) \right\}. \label{eq:LikeSBM}
\end{flalign}
Given the estimated $\widehat{\mathbf{g}}$, let $\widehat{N}_k = |\Gamma_k(\widehat{\mathbf{g}})| = |\{V_i:\widehat{g}_i=k\}|$ be the number of nodes assigned to community $k$. Define the number of possible edges between communities $k$ and $l$ as $n_{kl}=\widehat{N}_k \widehat{N}_l$ for $k \neq l$ and $n_{kk}=\widehat{N}_k(\widehat{N}_k-1)$ for $k = l$, and the number of observed edges for subject $r$ at time $t$ as $m_{kl}^{r,t} = \sum_{i<j} w_{ij}^{r,t} 1\{\hat{g}_i = k, \hat{g}_j = l\}$, where $1\{\cdot\}$ is an indicator function. We can re-write (\ref{eq:LikeSBM}) as
\begin{multline}
f({\bf W};\boldsymbol{\Theta},{\bf g}) = \sum_{r=1}^R \sum_{t=1}^T \sum_{k \leq l} \left\{ m_{kl}^{r,t} \log \theta_{kl}^{r,t} \right. \\
\left. + (n_{kl}-m_{kl}^{r,t}) \log (1-\theta_{kl}^{r,t}) \right\}. \notag
\end{multline}
Then ML estimate of connectivity parameters $\boldsymbol{\Theta}$ is given by
\begin{equation}
\hat{\theta}_{k,l}^{r,t} = \frac{m_{kl}^{r,t}}{n_{kl}}, \ \ t=1, \ldots, T, \ r=1, \ldots, R, \ k,l = 1, \ldots, K. \label{eq:mle}
\end{equation}
The estimated inter-block connection probabilities $\hat{\theta}_{k,l}^{r,t}$ correspond to the ratios of number of observed edges $m_{kl}^{r,t}$ relative to possible edges $n_{kl}$ within each block, which are also called as block densities.

\vspace{-0.1in}
\subsection{Identification of Dynamic Community States}
We fit an HMM in (\ref{eq:HMM-state})-(\ref{eq:HMM-obs}) on the estimates $\widehat{\boldsymbol{\Theta}} = \lbrace \widehat{\boldsymbol{\Theta}}^{r,t}; t=1, \ldots, T,  r=1, \ldots, R\rbrace$ to identify the distinct, recurring states in the time-evolving inter-modular connectivity that are common across subjects. The timing of shifts between states and the modular connectivity pattern in each state can be estimated simultaneously. Let $(\widehat{\boldsymbol{\beta}}^{1,1},\widehat{\boldsymbol{\beta}}^{1,2}, \ldots,  \widehat{\boldsymbol{\beta}}^{R,T})$ be a set of $RT$ concatenated vectors of the logit of estimated time-varying modular connection probabilities over all subjects.
Given a set of HMM parameters $\boldsymbol{\lambda}= \{\mathbf{\Pi}, \boldsymbol{\mu}^{[m]}_{\Theta}, \boldsymbol{\Sigma}^{[m]}_{\Theta}\}$, the temporal dynamics of the states over subjects $\lbrace {s}_{r,t} \rbrace$ can be obtained by extracting the most likely state sequence using the Viterbi algorithm
\begin{equation}
\widehat{s}_{1,1},\ldots, \widehat{s}_{R,T} = \argmax_{s_{1,1}, \ldots, s_{R,T} } p(s_{1,1},\ldots, s_{R,T}, \widehat{\boldsymbol{\beta}}^{1,1}, \ldots,  \widehat{\boldsymbol{\beta}}^{R,T} \vert \boldsymbol{\lambda}). \label{eq:VB}
\end{equation}
The state-specific block connectivity parameters $\{\boldsymbol{\mu}^{[m]}_{\Theta}, \boldsymbol{\Sigma}^{[m]}_{\Theta}\}$ can be estimated via ML by using the Baum-Welch algorithm \cite{Rabiner1989}. One could also fit an HMM to each subject and compute group-level parameters by averaging subject-specific estimates. Bayesian inference via the Markov chain Monte Carlo sampling \cite{Scott2002} can be used in our framework as an alternative estimation approach for HMM, which allows incorporation of prior information to improve parameter estimates.

\vspace{-0.05in}
\section{Simulations}

In this section, we shall evaluate the performance of our method on synthetic multi-subject networks.

\subsubsection{Community Recovery}
In this simulation, we first access the performance of different community detection methods in recovering a consensus community partition that is shared across different subjects. We generate binary networks of $R$ subjects from the multilayer SBM with balanced community size (with $N$ nodes equally partitioned into $K$ communities). The true community labels of nodes $\mathbf{g}$ are fixed and common across subjects. The modular connectivity matrix is set using the parameterization in \cite{Jing2015}
\begin{equation}
\boldsymbol{\Theta}= \alpha\boldsymbol{\Theta}_0: \ \ \boldsymbol{\Theta}_0 = \lambda \mathbf{I}_K + (1-\lambda){\bf 1}_K{\bf 1}^T_K, \ \ 0<\lambda<1 \label{eq:theta}
\end{equation}
where $\mathbf{I}_K$ is the $K \times K$ identity matrix and ${\bf 1}_K$ is the $K \times 1$ vector of 1's. The quantity $\lambda$ reflects the relative difference of the within- and between-community edge probabilities. The network sparsity is controlled by $\alpha$, where $N \alpha$ provides an upper bound on the average expected node degree. It is more difficult to recover the communities when $\alpha$ and $\lambda$ are close to $0$. We also allow inter-subject variability in the connectivity matrix by adding some subject-specific random deviations such that $\boldsymbol{\Theta}^r = \boldsymbol{\Theta} + \epsilon_r \mathbf{I}_K$ with $\epsilon_r \sim U[-0.1, 0.1]$.

We compare the performance of the proposed multilayer modularity maximization ($\text{Q}_{\max}$) algorithm with two single-layer methods: (1) Spectral clustering which performs K-means clustering on the $K$ leading eigenvectors of the graph Laplacian \cite{Rohe2011}, and (2) Single-layer $\text{Q}_{\max}$ using the Louvain algorithm (in Section III.A.(1)) as baseline. Both competing methods are widely used for community recovery with promising empirical performance, and have been shown to enjoy good statistical guarantee under the SBM \cite{Zhao2012,Jing2015}. Note that the number of communities $K$ was assumed known for the spectral clustering, and estimated from the simulated data for the $\text{Q}_{\max}$ methods. We measure the performance of all methods by the adjusted Rand index (ARI) between the ground-truth community labels $\mathbf{g}$ and their estimates $\hat{\mathbf{g}}$. The ARI is a measure of similarity between two partitions, taking values between 0 (random label assignments) and 1 (perfect recovery of true partition).

\begin{figure}[!t]
\hspace{-0.55 cm}
\centering
	\begin{minipage}[t]{0.5\linewidth}
		\centering
		\subfloat[]{\includegraphics[width=1\linewidth,keepaspectratio]{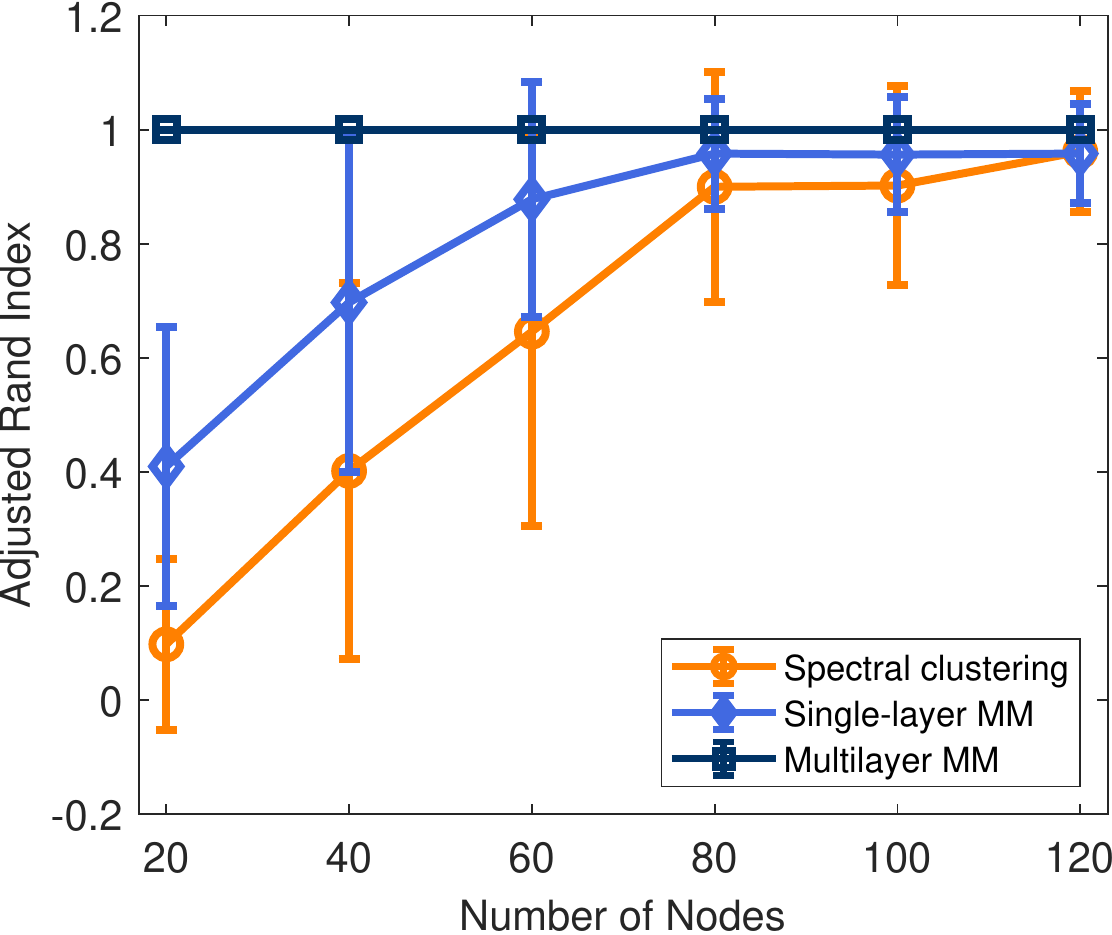}}
	\end{minipage}
	\hspace{-0.1cm}
	\begin{minipage}[t]{0.5\linewidth}
		\centering
		\subfloat[]{\includegraphics[width=1\linewidth,keepaspectratio]{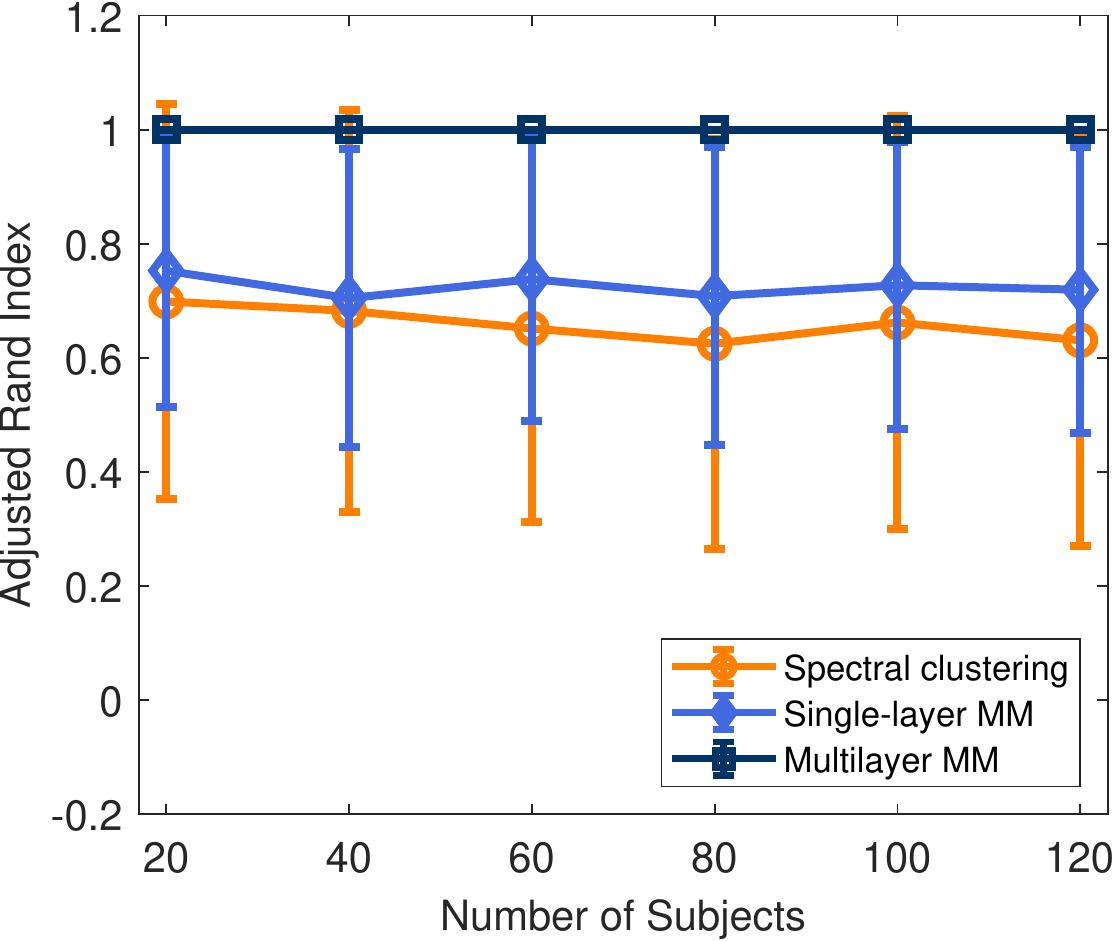}}
	\end{minipage} \\ \vspace{0.1 cm}
	\hspace{-0.55 cm}
		\begin{minipage}[t]{0.5\linewidth}
		\centering
		\subfloat[]{\includegraphics[width=1\linewidth,keepaspectratio]{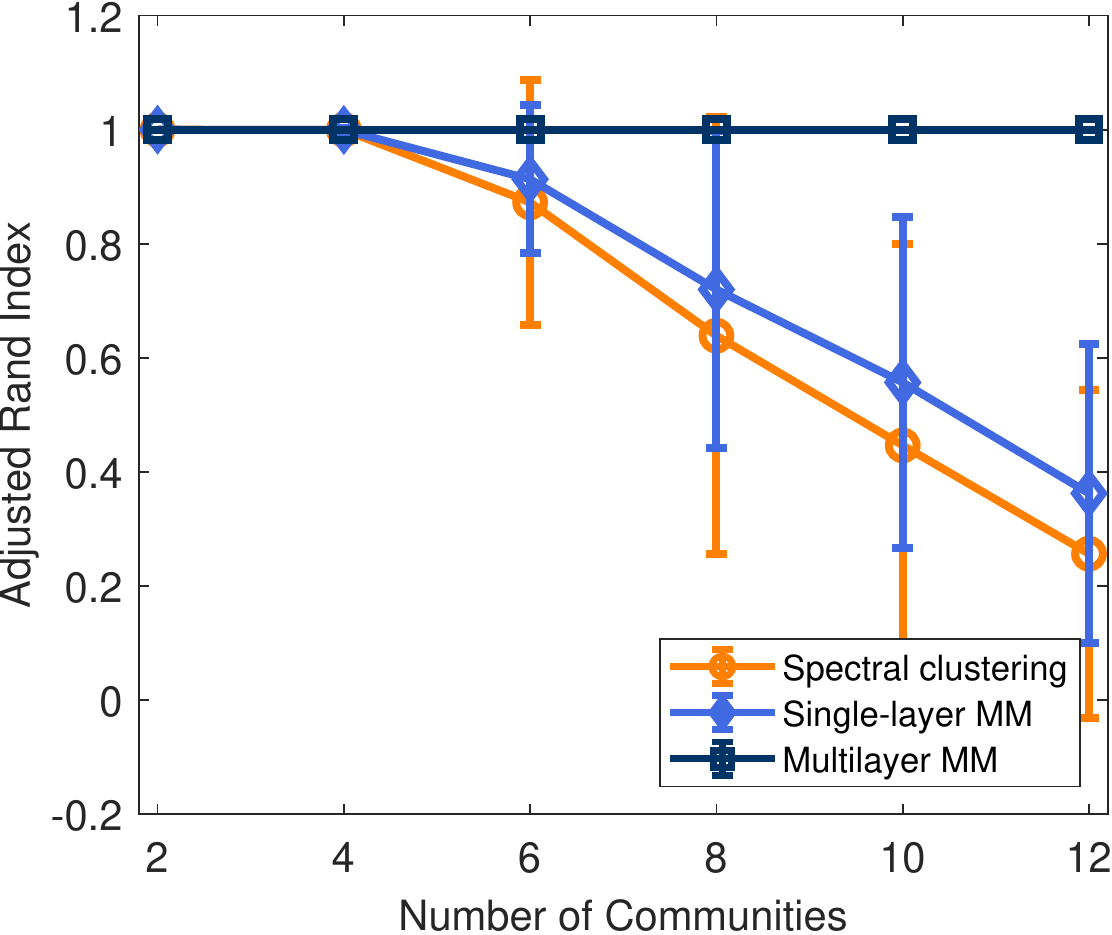}}
	\end{minipage}
	\hspace{-0.1cm}
	\begin{minipage}[t]{0.5\linewidth}
		\centering
		\subfloat[]{\includegraphics[width=1\linewidth,keepaspectratio]{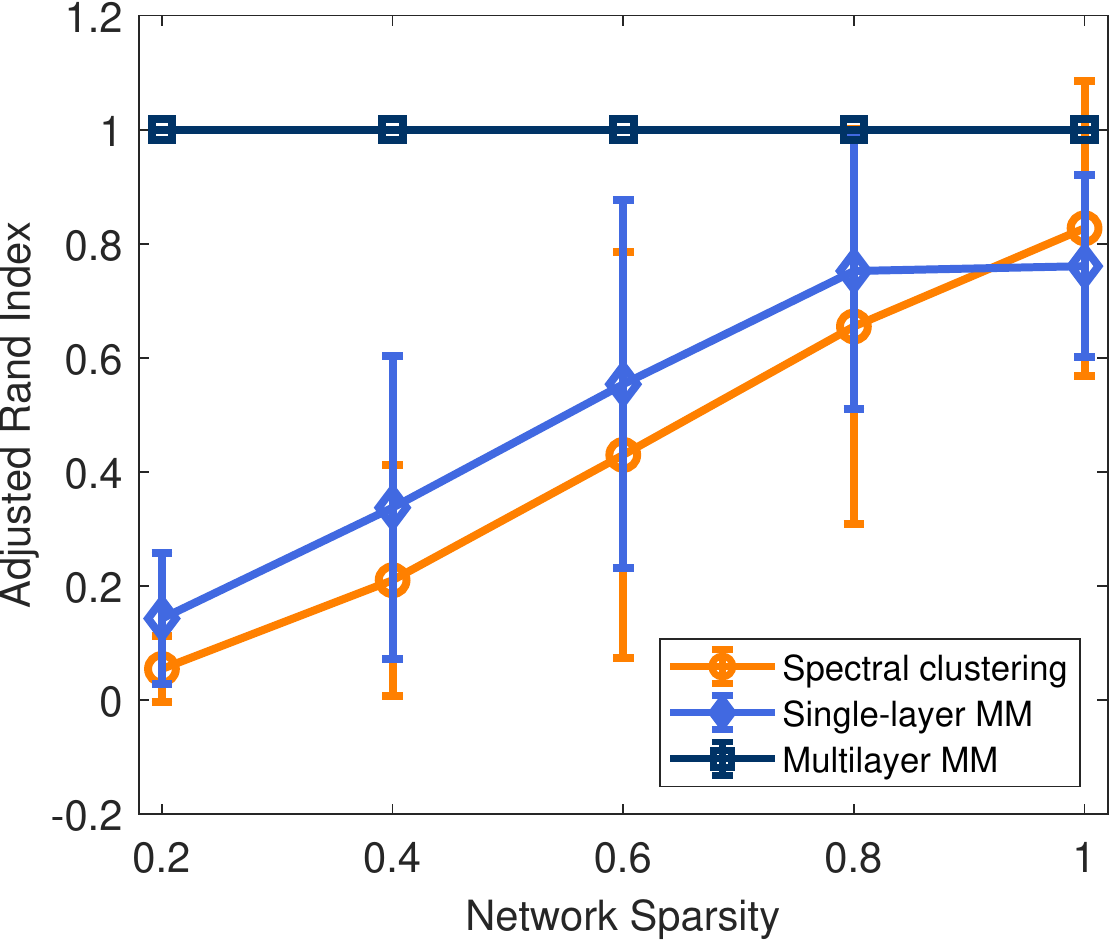}}
	\end{minipage}
\caption{Performance comparison of various methods for community detection in multisubject networks as measured by ARI between ground-truth and estimated community labels for four simulation settings. (a) Number of nodes $N$ increases, $K=5$, $R=100$, $\alpha=0.8$. (b) Number of subjects $R$ increases, $N=120$, $K=8$, $\alpha=0.8$. (c) Number of communities $K$ increases, $N=120$, $R=100$, $\alpha=0.8$. (d) Different levels of network sparsity $\alpha$, $N=120$, $K=8$ and $R=100$. Lines and error bars represent means and standard deviations over subjects.}
\label{Fig:SimCommRec}
\vspace{-0.1in}
\end{figure}

Fig.~\ref{Fig:SimCommRec} shows the performance comparison, in terms of ARIs over individual subjects, under different scenarios: increasing number of nodes $N$, number of subjects $R$, number of communities $K$ and varying levels of network sparsity $\alpha$. From Fig.~\ref{Fig:SimCommRec}(a), we see that the multilayer $\text{Q}_{\max}$ clearly outperforms the single-layer methods, achieving perfect recovery even when the number of nodes per community is very small. This suggests the robustness of the multilayer $\text{Q}_{\max}$ in small sample settings due to the pooling of data across multiple layers to estimate the common community structure accurately. The performance of both single-layer methods improves steadily as $N$ increases, with the single-layer $\text{Q}_{\max}$ reaching ARI of close to 1 faster than the spectral clustering. Fig.~\ref{Fig:SimCommRec}(b) shows that small number of layers/subjects ($R=20$) is sufficient for the multilayer $\text{Q}_{\max}$ to yield an exact community reconstruction. The single-layer methods have slightly lower accuracy of community detection and do not show any improvement with more layers because they carry out the community detection in individual networks independently. Fig.~\ref{Fig:SimCommRec}(c) shows that the multilayer $\text{Q}_{\max}$ is able to consistently recover the communities for large number of communities. The ARI of both single-layer methods drops with increasing $K$, at a much faster rate for the spectral clustering compared to the single-layer $\text{Q}_{\max}$. In Fig.~\ref{Fig:SimCommRec}(d), the single-layer methods especially the spectral clustering perform poorly when the network is sparse (low values of $\alpha$). This agrees with other studies which have shown that spectral methods tend to suffer from inconsistency in sparse graphs \cite{Krzakala2013}. As expected, the accuracy of these methods increases when the networks become denser as $\alpha$ increases. In contrast, the multilayer $\text{Q}_{\max}$ remains robust even in the sparse network case. Additional simulation shows the multilayer $\text{Q}_{\max}$ algorithm converges faster than single-layer $\text{Q}_{\max}$ (See Supplementary Section 4).

\subsubsection{Estimation of State-Related Changes}

We further evaluate the proposed MSS-SBM in identifying underlying temporal regime changes in the modular connectivity patterns. We generated time series of synthetic dynamic functional networks for a cohort of $R$ subjects, according to the time-varying, multi-subject SBM of (\ref{Bernoulli-edge}). To emulate the hidden dynamics of regime switching in the network community structure, the simulated sequences of $T=240$ temporal graphs $\{\mathbf{W}^{r,t}\}$ were characterized by time-evolving modular connectivity $\boldsymbol{\Theta}^{[s_t]}$ driven by underlying piece-wise stationary state time course $s_t \in 1, \ldots, S$. Here we drop the subject index $r$ in $s_t$ assuming all subjects to share identical state dynamics. To imitate the typical block-design paradigm in task-based fMRI experiment, time-blocks of states (each state represents a task or stimulus) were interleaved and repeated over the time course. This reproduces quasi-stable, recurring network modular structure over time points. We consider $S=3$ distinct states each has a unique modular connection probability matrix $\boldsymbol{\Theta}^{[m]}$ of the form (\ref{eq:theta}) with $\lambda = 0.9$ for $m=1$, $\lambda=0.75$ for $m=2$ and $\lambda = 0.6$ for $m=3$, which represent states of high, medium, and low within-community connectivity, respectively. We further introduce temporal variability by adding random fluctuations into the piecewise constant trajectory $\boldsymbol{\Theta}^{[s_t]}$, i.e., $\boldsymbol{\beta}^{t} = \boldsymbol{\beta}^{[s_t]} + \boldsymbol{\eta}^{t}$, $\boldsymbol{\eta}^{t} \sim N(\bf{0},\sigma \bf{I})$ where $\boldsymbol{\beta}^{[s_t]} = \text{vec}(g(\boldsymbol{\Theta}^{[s_t]}))$. The estimate of state sequence $\hat{s}_t$ is obtained via Viterbi algorithm, and the state-specific connectivity parameters $\widehat{\boldsymbol{\Theta}}^{[m]} = E[\boldsymbol{\Theta}^{t}|s_t = m]$ from the inverse logit of the estimated means of HMM Gaussian observation density $g^{-1}(\widehat{\boldsymbol{\mu}}^{[m]}_{\Theta})$.

To measure dynamic state estimation performance, we compute the ARI between the true and estimated partitions of the $T$ temporal observations into states based on $\widehat{s}_t$. It is defined in terms of numbers of pairs of time points that are correctly identified as belonging to the same or different states, where ARI = 1 indicates perfect recovery of the true state sequence. This measure also indirectly evaluates the change-point detection in the modular connectivity structure. We also calculate the mean squared error between the ground-truth and estimated time-evolving modular connectivity matrices over the time course, MSE$= T^{-1} \sum_{t=1}^T \vert\vert \widehat{\boldsymbol{\Theta}}^{[\widehat{s}_t]} - \boldsymbol{\Theta}^{[s_t]} \vert\vert^2_F$ where $\vert\vert \mathbf{H} \vert\vert_F = tr (\mathbf{H}'\mathbf{H})^{1/2}$ denotes Frobenius norm of matrix $\mathbf{H}$.

\begin{figure}[!t]
\hspace{-0.55 cm}
\centering
	\begin{minipage}[t]{0.5\linewidth}
		\centering
		\subfloat[]{\includegraphics[width=1\linewidth,keepaspectratio]{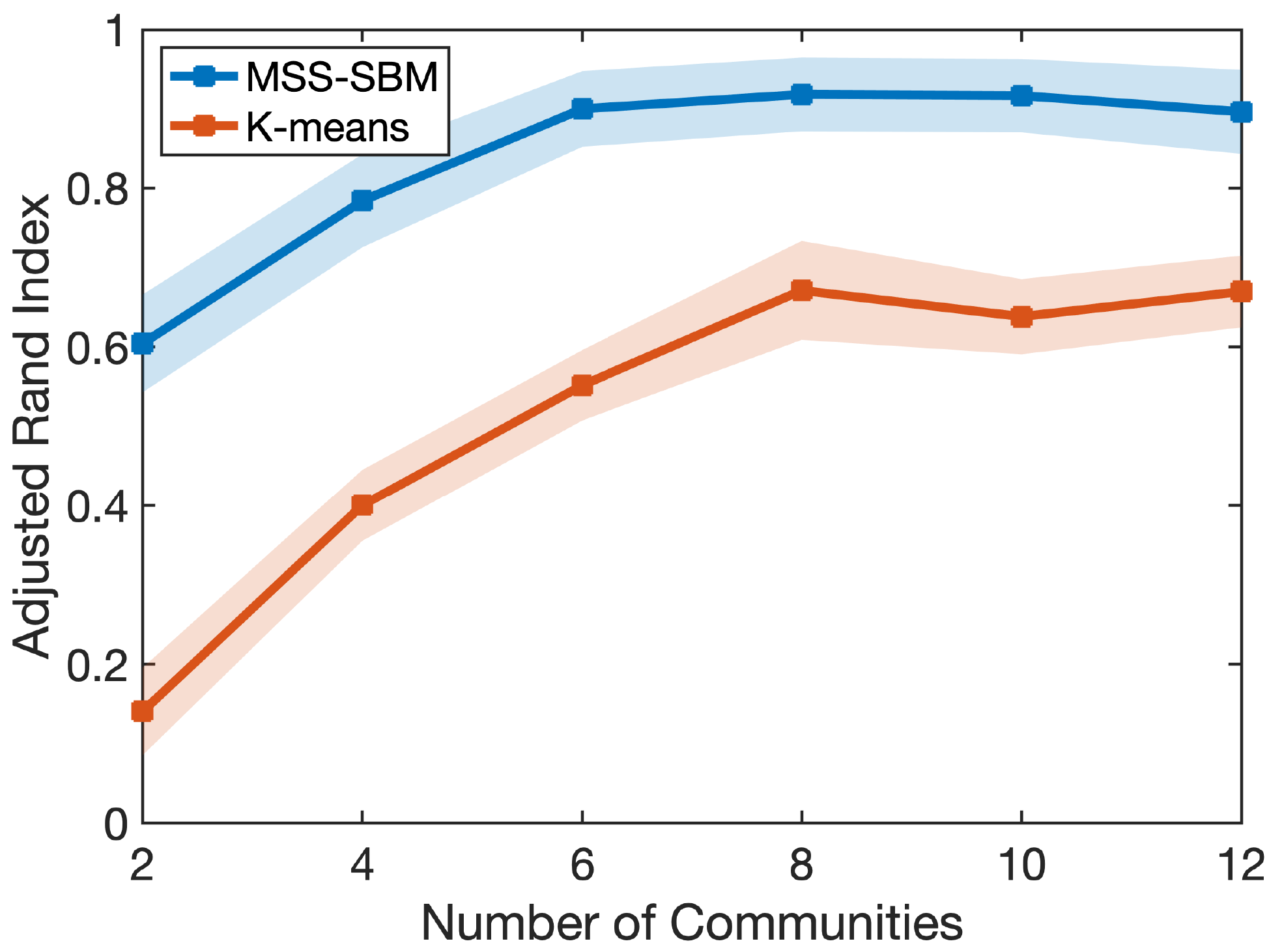}}
	\end{minipage}
	\hspace{-0.1cm}
	\begin{minipage}[t]{0.5\linewidth}
		\centering
		\subfloat[]{\includegraphics[width=1\linewidth,keepaspectratio]{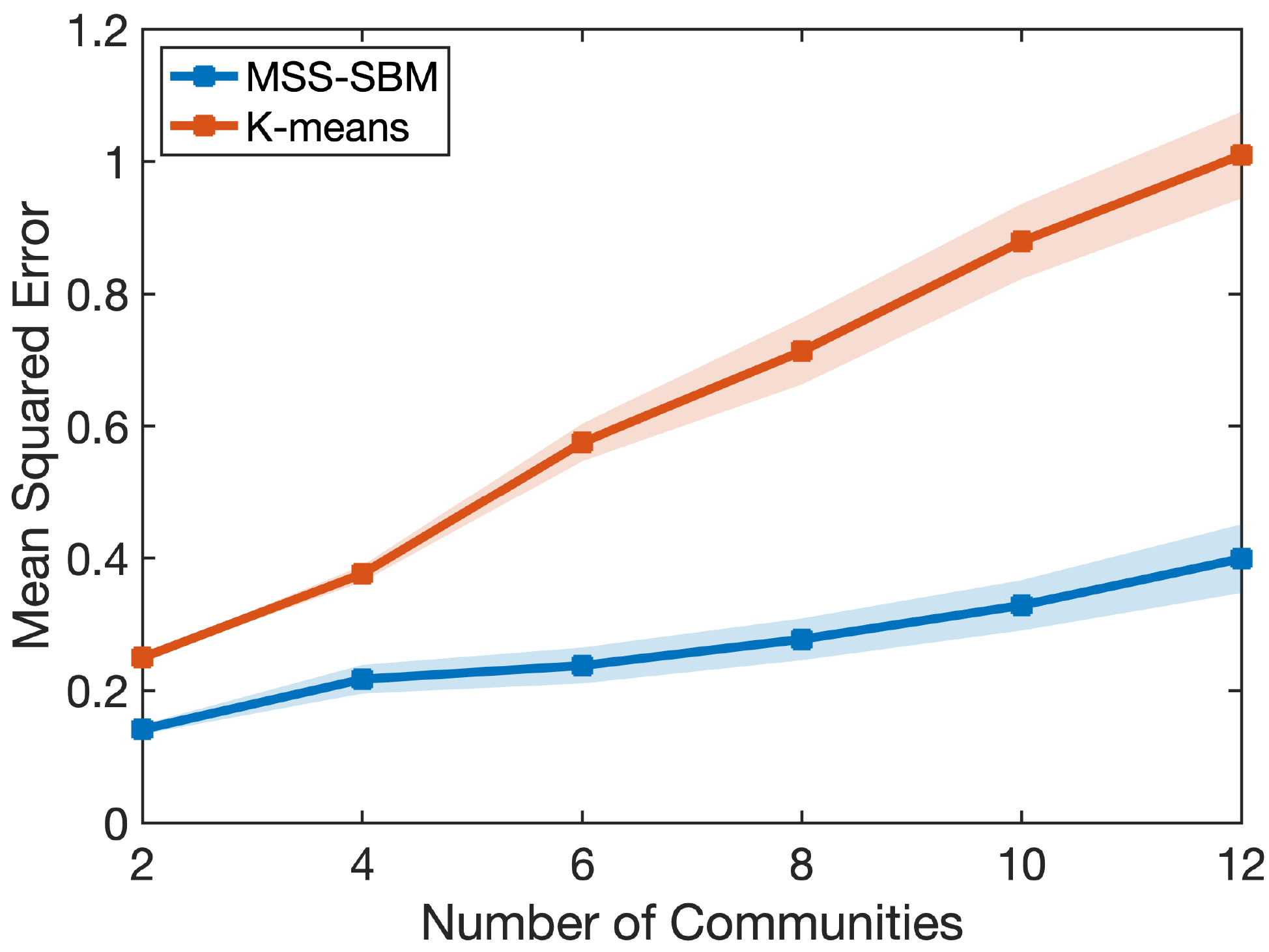}}
	\end{minipage}
\caption{Performance of MSS-SBM and K-means clustering in estimating state-based dynamic inter-modular connectivity in simulated multi-subject networks for increased number of communities $K$ with $N=120$, $T=240$ and $S=3$. (a) ARI of dynamic state identification. (b) MSE for estimated time-dependent connectivity matrices. The intervals represent standard deviations over $R=50$ subjects and 100 replications of simulation for each subject.}
\label{Fig:SimState}
\vspace{-0.15in}
\end{figure}

We access the scalability of MSS-SBM for dynamic connectivity estimation in the presence of large number of communities $K$. Fig.~\ref{Fig:SimState} plots the ARIs and MSEs over $R=50$ subjects as a function of increasing $K$ with fixed $N=120$. Compared to the K-means clustering approach which is widely used in estimating dynamic brain connectivity states, the MSS-SBM performs better in both tracking of dynamic regimes and connectivity estimation. This implies the advantages of using Markov process in the MSS-SBM to model the temporal evolution of the connectivity states, and the Gaussian observation density to account for variations within each state. Fig.~\ref{Fig:SimState}(a) shows improved accuracy of dynamic regime identification with increasing $K$. This is because larger $K$ may provide more information about the distinct inter-community connectivity structure that allows better discrimination between different temporal states. Despite improved temporal state partitioning, estimation errors of connectivity matrices increase as $K$ increases (Fig.~\ref{Fig:SimState}(b)). This is due to larger number of parameters in the state-specific connectivity matrices to be estimated, relative to fixed sample size available for each state.

\vspace{-0.05in}
\section{Application to Task fMRI}

We examined the task-based dynamic FC, a subject of intensive recent research to understand how brain networks reconfigure dynamically to accommodate task demands \cite{Gonzalez2018}. Most FC studies focus on resting-state fMRI which suffers drawbacks due to its unconstrained nature, e.g., dynamic FC states mapped during rest fail to provide a clear link to ongoing cognitive states, where the identified network dynamics might be driven by artifactual rather than functionally-relevant sources \cite{Mill2017}. In task settings, cognitive relevance of such FC states can be evaluated, at least with respect to externally-imposed tasks, using the ``ground-truth" defined by timing of task performance in the experimental paradigm (e.g., to which task a given temporal segment belongs).

Specifically, we apply the proposed MSS-SBM approach to identify state-driven dynamic switching in inter-community interactions of fMRI functional networks across subjects as modulated by distinct task performance. We assume there is a common community partition across subjects and time, but allow the inter-modular connectivity to vary as evoked by alternating conditions over the time course of experiment. The number of connectivity states corresponds to the number of conditions, however the timing of changes in the connectivity states between conditions is unknown a priori. Based on adjacency matrices estimated from fMRI data, we evaluated both the subject-specific and multi-subject community detection using the single-layer and multilayer $\text{Q}_{\max}$ algorithm, respectively. Given the community partition, time-evolving inter-modular connectivity matrices were estimated by ML followed by HMM fitting to identify the dynamic connectivity states. We analyzed task-related fMRI data of 400 subjects for language tasks and 450 subjects for motor tasks from the Human Connectome Project. See Supplementary Section 1 for details of experimental design, data acquisition and preprocessing.

\vspace{-0.05in}
\subsection{Network Construction}

To estimate the dynamic FC, we computed Pearson's correlations between the 90 ROI time series over sliding windows of 30 time points / TRs ($\sim22$ s for HCP data) with a step size of 1 TR. See Supplementary Section 3.2 on detailed validation analysis for the proposed choice of window length. For the construction of time-varying functional networks, the sliding-window correlation matrices were thresholded to create time-varying adjacency matrices. The thresholding of FC matrix defines the edges in the adjacency matrix and therefore has a direct effect on the subsequent computation of graph metrics including the modularity. Here we used the proportional thresholding \cite{Schmidt2017} by setting a fraction $\kappa$ of strongest connections (with the highest absolute correlation values) of the derived FC matrix for each individual network to 1, and other connections to zero. The application of the proportional threshold $\kappa$ will result in a binary graph with connection density of $\kappa$, defined for undirected graph as $\kappa = 2 \epsilon/ N(N-1) $, $\kappa \in (0,1)$ where $\epsilon$ is the number of preserved edges \cite{Bullmore2011}. This approach will produce a fixed density of edges in graphs across all subjects and time windows, and thus enabling meaningful comparison of network topology between different groups and conditions. It has been shown to generate more stable network metrics compared to the absolute thresholding \cite{Garrison2015}. By exploring different topological properties of the resulting networks over a range of connection densities, the threshold $\kappa=0.25$ was identified as optimal indicating a balance between network segregation and integration (See Supplementary Section 3.1).

\vspace{-0.05in}
\subsection{Results for Subject-Specific Community Detection}

To detect subject-specific community structure, we first computed subject-specific adjacency matrices for static connectivity by proportional thresholding the time-averaged correlation matrices of individual subjects. By applying the Louvain algorithm, the detected number of communities varies across different subjects (Supplementary Fig.~2.1). To obtain a consistent mapping of community partition across subjects from single-subject analysis is non-trivial since each subject has different number of communities. To solve this, we computed the community association matrices to quantify the occurrence that pairs of nodes belong to the same community (Supplementary Fig.~2.2), and aggregated them over all subjects to obtain a consensus community partition (Supplementary Fig.~2.3).

We then analyzed the regime changes in the community structure of dynamic functional networks using Markov-switching SBM (MS-SBM) as in \cite{Balqis2019} - a special case of MSS-SBM for single-subject analysis. We fitted MS-SBM with $S=2$ and $S=6$ states on the time-varying adjacency matrices for the language and motor tasks, respectively. Here we assume the number of dynamic community states $S$ corresponds to the number of tasks in the experiments. We present a data-driven procedure to estimate $S$ when it is unknown \textit{a priori} (e.g., for the resting-state), via clustering analysis of $\widehat{\boldsymbol{\Theta}}^{r,t}$ (See Supplementary Section 3.3). Using silhouette and Davies-Bouldin cluster validity indices, the selected $S$ on both task fMRI was close to the number of tasks (Supplementary Table 3.1). Given the estimated subject-level community membership of the 90 ROIs using Louvain algorithm, we computed the subject-specific, time-varying inter-modular connectivity parameters by ML method as in (\ref{eq:mle}) and then fitted an HMM for each subject individually to detect the dynamic community regimes. The performance of Viterbi algorithm in (\ref{eq:VB}) in tracking temporal regimes is compared with the K-means clustering which is widely used in dynamic connectivity state estimation.

\begin{figure}[!t]
\centering
	\includegraphics[width=0.97\linewidth,keepaspectratio]{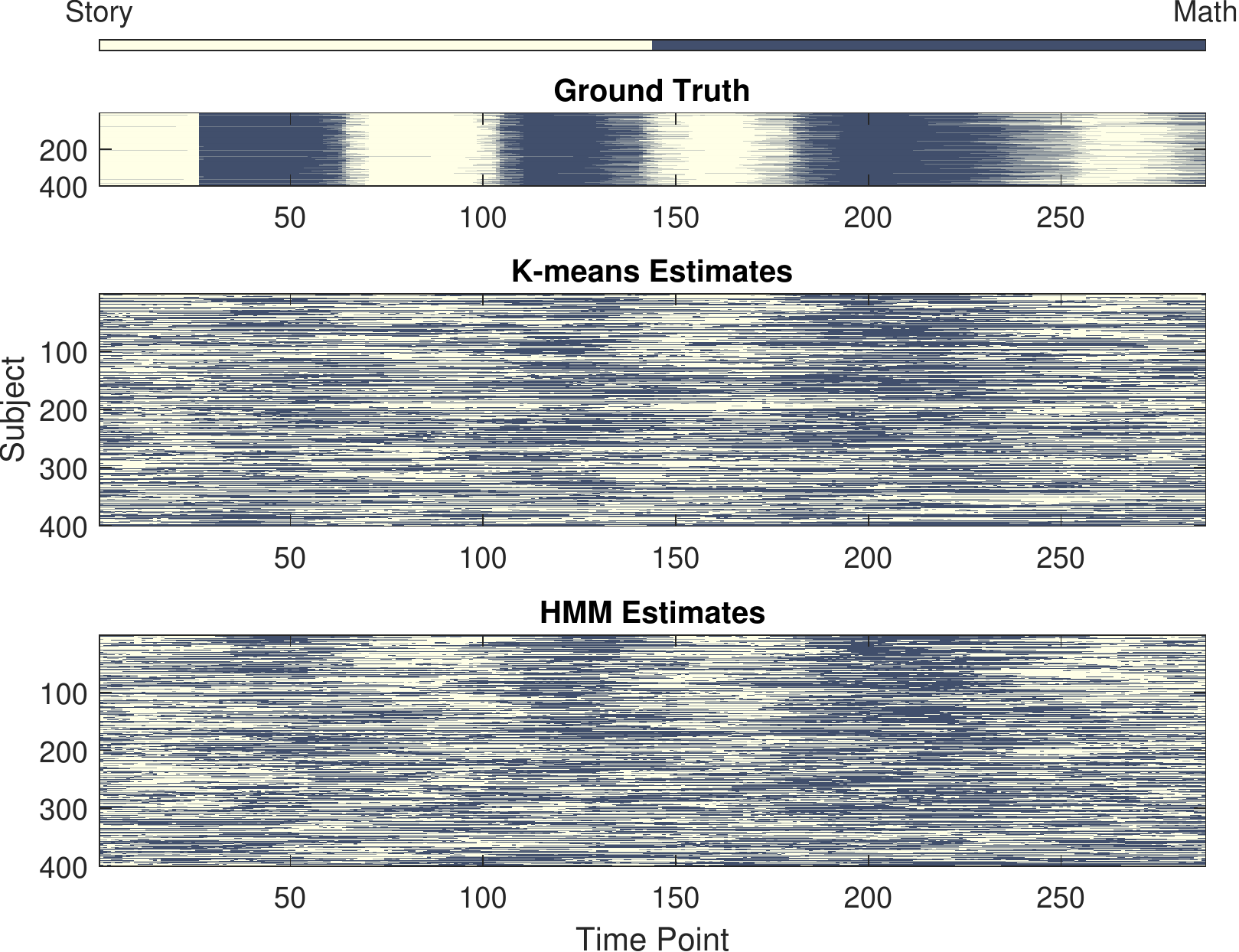}
	\vspace{-0.12in}
\caption{Tracking of regime changes in the inter-community connectivity of fMRI functional networks over individual subjects for language task experiments. See Supplementary Fig.~2.4 for motor task. (Top) Ground-truth task state time courses from experimental design. (Middle \& Bottom) Dynamic state estimates by K-means clustering and MS-SBM.} \label{LM-TemStateAcc}
\vspace{-0.2in}
\end{figure}

Fig.~\ref{LM-TemStateAcc} shows the tracking of dynamic regimes of inter-modular connectivity for each subject via the estimated state sequence $\lbrace \widehat{s}_{r,t} \rbrace$ for the language task. Compared to K-means approach, the estimates by MS-SBM show better tracking of temporal regimes changes in modular connectivity, which follow more closely the changes in task conditions (indicated by the ground-truth state sequence) over the time course of experiment. The MS-SBM provides a more accurate detection of abrupt change points between regimes than the K-means clustering which produces more spurious temporal state estimation. Similar better performance in identifying dynamic states was observed for the motor tasks (Supplementary Fig.~2.4). Note that the state estimation by MS-SBM was accomplished in an unsupervised manner, i.e., without being pre-trained from labeled data. Despite the good performance of MS-SBM, the aim is not an exact recovery of task states from experimental designs but to investigate the inter-subject variability in the dynamic community structure of brain networks. We can see considerable heterogeneity in the temporal state dynamics across subjects, probably due to individual differences in response to changes in tasks and stimuli.

\begin{figure}[!t]
\centering
\includegraphics[width=0.9\linewidth,keepaspectratio]{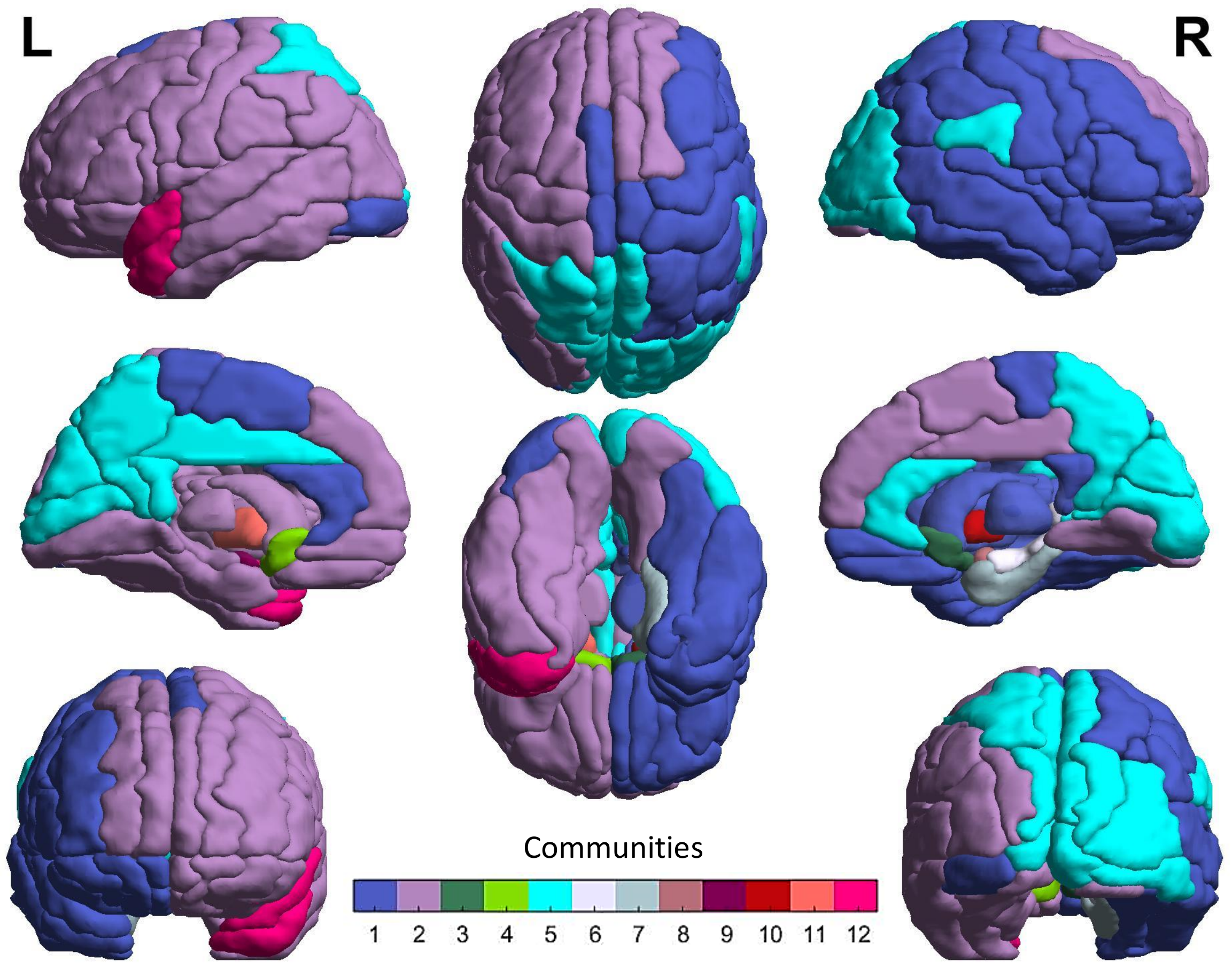} 
\vspace{-0.1in}
		\caption{Topographic representation of group-level brain community partition detected over 400 subjects from the language task fMRI based on multilayer modularity maximization. The 90 ROIs were color-coded according to their assigned communities. See Supplementary Fig.~2.5 for motor tasks.}\label{LangCommMulti}
	\vspace{-0.2in}
\end{figure}

\vspace{-0.1in}

\subsection{Results for Multi-Subject Community Detection}

We applied the multilayer $\text{Q}_{\max}$ approach in Section III.A.2 to identify a group-level community structure of functional networks across subjects under the proposed MSS-SBM model. We averaged the time-varying connectivity matrices for individual subjects to construct subject-specific networks, which were concatenated to form a multilayer network ensemble in which layers represent single-subject networks. The nodes' community assignment in all subjects were then estimated simultaneously by maximizing the multilayer modularity in (\ref{eq:MSQ-function}) over all subjects, which was accomplished by the generalized Louvain algorithm \cite{Jeub2017}. We set $\gamma_r = \gamma = 1$ as commonly used for community detection for human brain networks \cite{Bassett2013b}. To determine the interlayer connection weight, we investigated a range of values $[0.9, 1, 1.1]$, and chose $C_{jrs} = C = 1$ which yields a consensus community partition across all subjects. Fig.~\ref{LangCommMulti} shows the group-level community assignment of the 90 ROIs for the language task over the 400 subjects. There are 12 communities detected, including 3 large communities (Community \#1, \#2 and \#5), one small community (\#12) and 8 singleton communities which are composed of a single node (\#3, \#4 and \#6 - \#11). The ROI members of the four largest communities are given in Supplementary Table 2.1. Almost 87\% of ROIs or nodes assigned to community \#1 are located in the left hemisphere of the brain and 86\% of nodes assigned to community \#2 in the right hemisphere. This is consistent with the well-established notion of functional lateralization of the human brain, particularly the superiority of the left hemisphere in language processing \cite{frost1999}. Recent fMRI studies also show the essential role of the right side of the brain in language \cite{Muller2014}, which explains the formation of large community in the right hemisphere. Community \#5 consists of more diverse nodes from both hemisphere suggesting its involvement in cross-hemisphere interaction. Interestingly, it comprises the key regions of language networks including middle-temporal-gyrus (MTG, both left and right), a major area involved in language processing, both comprehension and production. Community \#5 also includes the thalamus (THA, both left and right) which plays a central role in synchronizing separate areas within linguistic processing \cite{johnson2000,klostermann2013}. The detected community structure for motor task (Supplementary Fig.~2.5) mainly comprises motor and sensory areas (See Supplementary Table.~2.2).

\begin{figure}[!t]
\centering
	\includegraphics[width=0.9\linewidth,keepaspectratio]{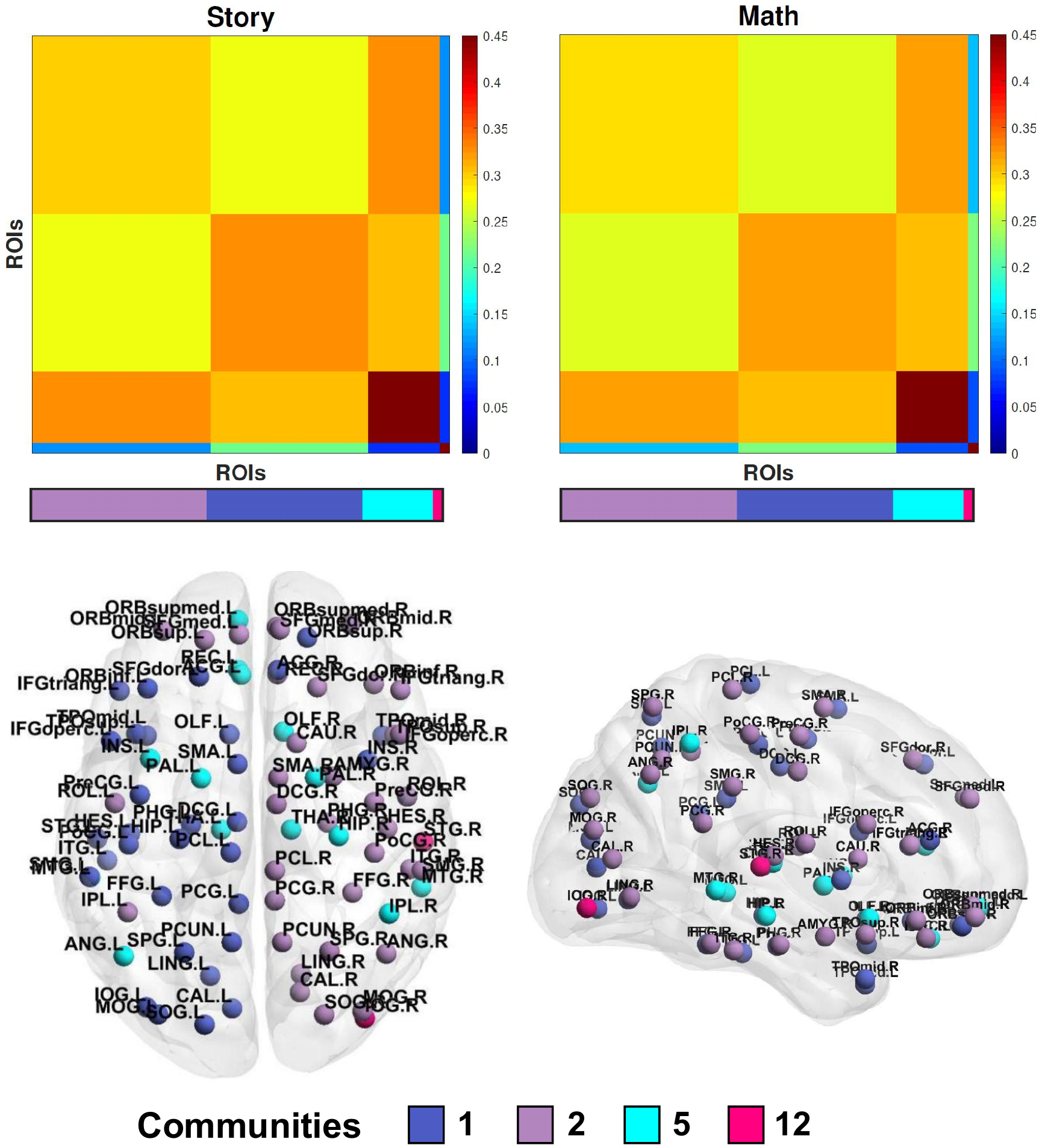}
	\vspace{-0.1in}
\caption{Estimated median modular connection probability matrices between four largest communities for two states in the language tasks: story (left) and math (right). The spatial distributions of the member ROIs of the communities are depicted in brain renders. See Supplementary Fig.~2.6 for the motor tasks.} \label{LM-Densitymat}
\vspace{-0.2in}
\end{figure}

\begin{figure}[!t]
\centering
\includegraphics[width=1\linewidth,keepaspectratio]{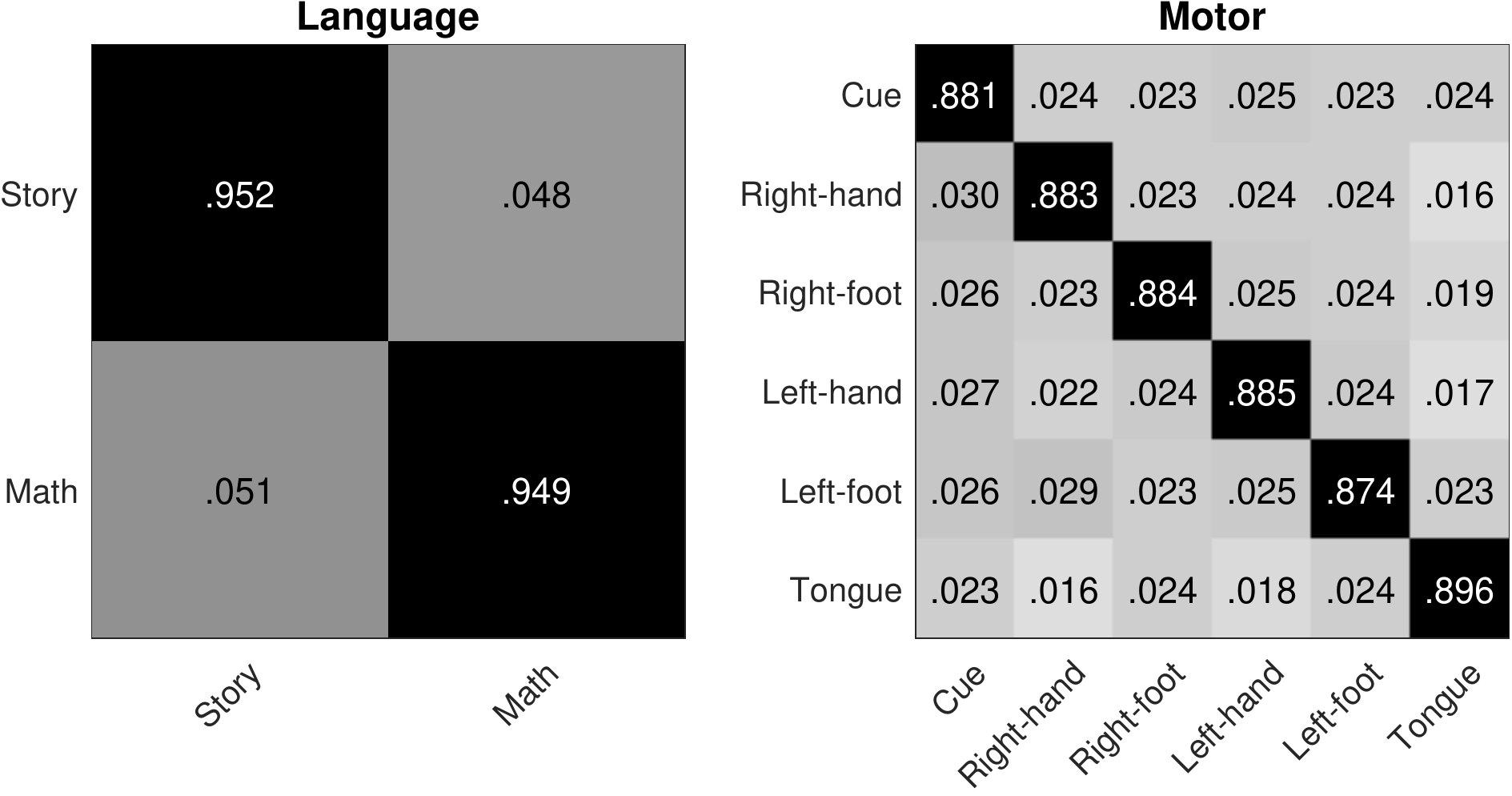}
\vspace{-0.3in}
\caption{Estimated transition probability matrices between states of the language and motor tasks.}  \label{transmat}
\vspace{-0.22in}
\end{figure}

Given the estimated common community structure, we computed the time-dependent inter-modular connection probabilities (higher probability indicates higher density of edges), and estimated state-based changes in connectivity patterns between communities using HMMs. Fig.~\ref{LM-Densitymat} shows the estimated group-level block connection probability matrices between four largest communities for story and math states of language task. Results are medians computed over subject-level estimates. We can see the networks exhibit non-purely assortative community structure mixed of assortative and core-periphery configurations. The communities of left and right hemispheres (\#1 and \#2) are assortative with denser connections within community of each hemisphere but sparser cross-hemisphere connections. These two segregated communities may engage in specialized information processing in language perception. Community \#5 with nodes from both hemispheres is non-assortative in the form of core-periphery motif. It acts as a core-like community with strong intra-community connection density while projecting inter-community interactions with periphery-like communities (\#1 and \#2) with relatively sparsely connected nodes. This may suggest its role of information integration in language processing, transiently broadcasting information to or receive information from periphery across hemispheres. The story and math tasks elicit similar connectivity patterns between large communities, partly due to the overlapping sensory and cognitive effort required in both tasks such as auditory and phonetic perception, syntactic analysis, attention and working memory. Nevertheless, slightly denser connection was detected between community \#1 and communities \#2 and \#5 in story task. This enhanced network integration may be essential to facilitate more complex semantic processing when comprehending spoken narratives, compared to the non-semantic processes in the math task. Non-assortative community motif was also found for all the six states of the motor tasks, as shown in the estimated modular connectivity matrices (Supplementary Fig.~2.6). The temporal evolution of the connectivity states is captured by the estimated transition probability matrices in Fig.~\ref{transmat}. The results indicate persistence in dwelling in the same states and occasional switching to other states (high self-transition and low inter-state transition probabilities), which is consistent with the block-design experimental paradigm used in the language and motor tasks.

\begin{figure*}[!th]
\centering
\includegraphics[width=0.75\linewidth,keepaspectratio]{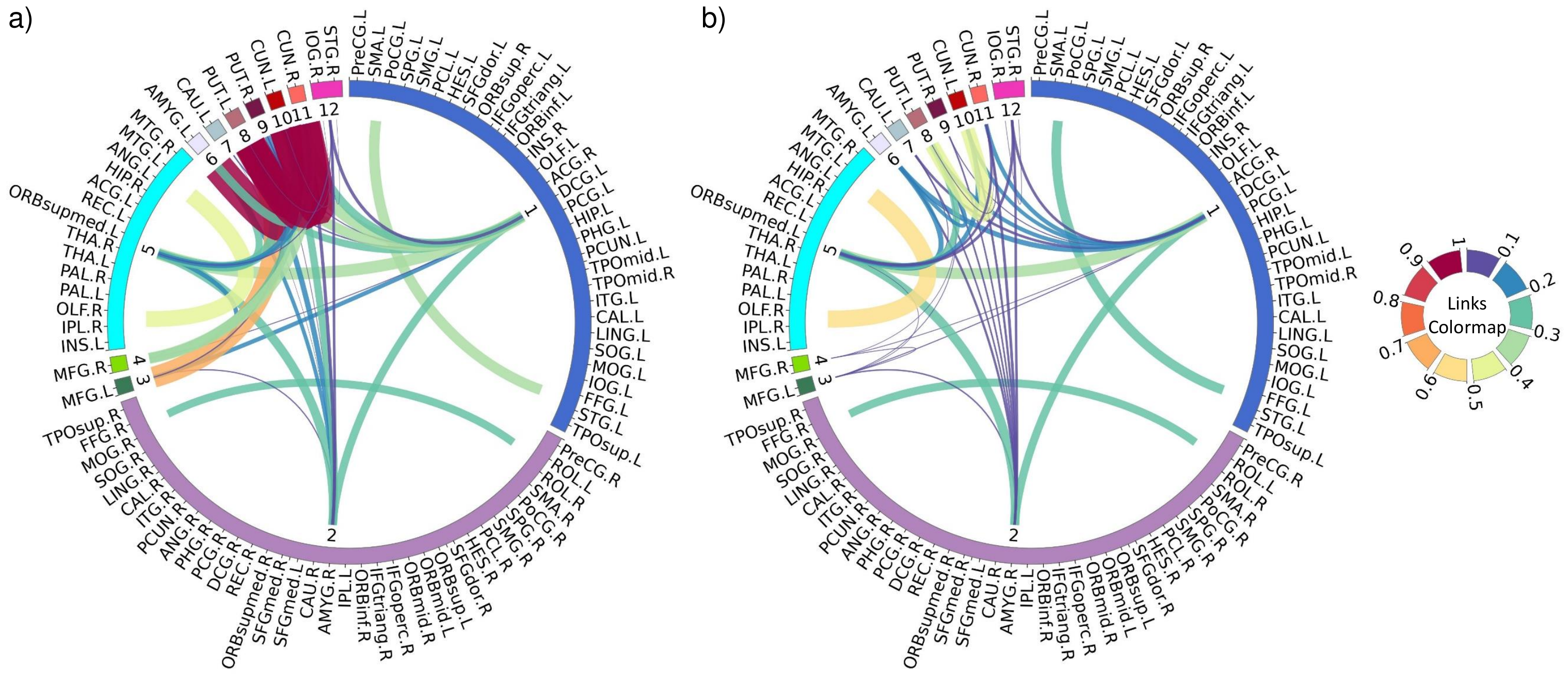} 
\vspace{-0.1in}
	\caption{Within and between-community connectivity for each state of the language task. (a) Story. (b) Math. See Supplementary Fig.~2.7 for motor tasks.} \label{LangMap}
	\vspace{-0.2in}
\end{figure*}

The difference in the inter-community connectivity patterns across distinct states is more pronounced when including the singleton communities. Fig.~\ref{LangMap} show interactions between all communities detected for the language task for a subject. The links represent the inter-block connection probabilities. We observed markedly distinct motifs of community interactions between the two states. We see stronger between-community connectivity in the story state than the math state. In the motor task, we also found unique network configuration within and between communities across the six different conditions (Supplementary Fig.~2.7).

\begin{figure}[!t]
\centering
\includegraphics[width=0.85\linewidth,keepaspectratio]{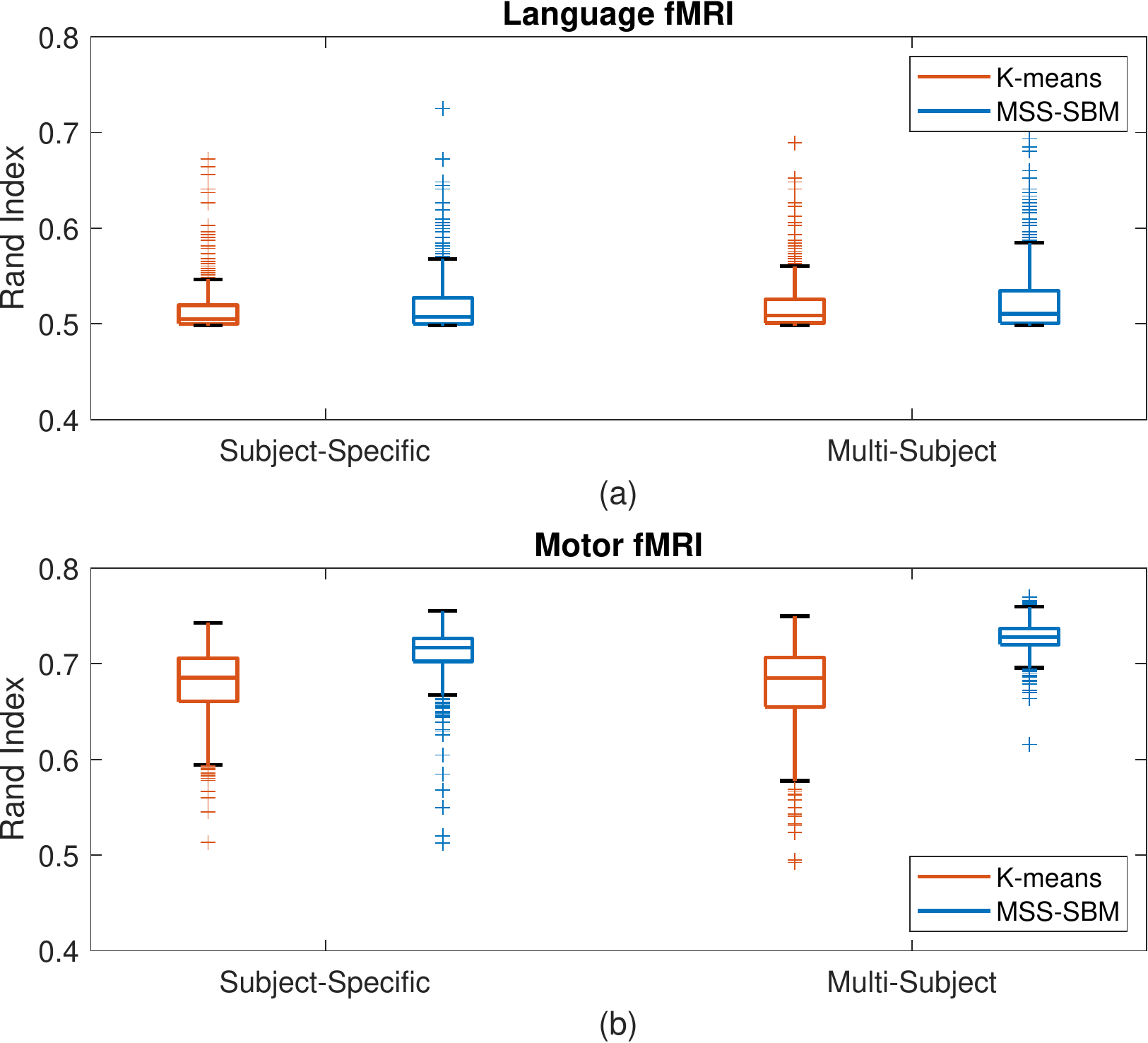} 
\vspace{-0.15in}
	\caption{Comparison of different community detection methods and temporal clustering methods for tracking modular connectivity state dynamics in language and motor task fMRI, as measured by RI values of estimated state time courses over subjects relative to the experimental ground-truth.} \label{RandIn_fmri}
\vspace{-0.2in}
\end{figure}

We evaluate dynamic state estimation performance under subject-level and group-level community partitions, detected by the subject-specific and multi-subject $\text{Q}_{\max}$ algorithms, respectively. We also compare HMM and K-means clustering in tracking dynamic connectivity states. Fig.~\ref{RandIn_fmri} plots distributions of RIs of the estimated state sequences over subjects relative to changes of conditions in the experiments. The higher RIs for both tasks confirm the results in Fig.~\ref{LM-TemStateAcc} on the superiority of MSS-SBM over the K-means clustering for detecting shifts between distinct states of modular connectivity. These results are supported by additional performance metric based on the $F$-measure (See Supplementary Section 2.4). Despite the advantage of subject-level community detection to account for inter-subject variability with varying numbers of communities and community organization for individual subjects, the use of common group-level community partition produces better results in RIs for the dynamic state estimation. This implies existence of shared community structure among subjects and synchronous brain dynamics in response to the same tasks or stimuli, which may not apply to resting-state data. We observe higher RIs for the motor task, reflecting the better alignment of the estimated state sequences with the ground-truth (Supplementary Fig.~2.4) compared to the language task (Fig.~\ref{LM-TemStateAcc}). This is likely due to the more distinct modular connectivity patterns across different states in the motor task (Supplementary Fig.~2.7) than the language task, which renders states in the motor task easier to be differentiated over the temporal dimension. For computational time of different algorithms see Supplementary Section 5.

\vspace{-0.12in}
\section{Discussion}

We developed a novel statistical framework based on a multilayer, Markov-switching SBM for identifying state-driven dynamic modular connectivity in multi-subject brain functional networks. We first propose a multilayer SBM, a generalization of existing dynamic SBM for single networks to an ensemble of networks, which provides a principled way of characterizing time-dependent inter-community connectivity of brain networks for a group of subjects. The model allows brain nodes to share common community partition over multiple network layers formed by aggregating connectivity matrices of individual subjects, but the inter-community connection density may vary flexibly across layers (subjects and time). By augmenting the multilayer SBM with a Markov-switching model to describe the temporal dynamics, it enables us to identify distinct, repeating states with respect to inter-community connectivity over time, without a priori assumption on the temporal locations of the transition between states. We further introduce the use of multilayer modularity maximization for estimating the latent block structure of the proposed MSS-SBM, which can uncover common community assignments in functional networks of many subjects simultaneously. This overcomes the problem of inconsistent mapping of community labels across subjects in the traditional subject-specific community detection.

Simulation results show the effectiveness of the proposed multilayer modularity maximization for recovery of common community structure in multilayer networks even when the network is sparse and the number of communities to be detected is large. When applied to two sets of HCP task-based fMRI data, our method detected more diverse community organization in addition to the typical assortative structure in brain networks, which is associated with language processing and motor functions. Even more notable is that our method was able to identify non-assortative community motifs such as the core-periphery structure. These types of network architecture engender more complex inter-community interactions that may allow the network to engage in a wider functional repertoire, e.g., integration of information across different brain regions in higher-order cognitive processes. For example, we found a bilateral core-like community in the language network that subserves an integrative function between periphery communities in the left and the right hemisphere during language comprehension. The proposed MSS-SBM also captures state-related dynamic re-configuration of inter-modular interactions in the brain networks, as modulated by the repetitive changes in task conditions over time course of experiment. It identified a set of putative network states with distinct profiles of within and between-community connectivity that are differential between task conditions, such as left and right movements in the motor fMRI data.
Our method has produced findings that could lead to new sets of hypotheses about dynamic brain functional networks particularly the state-driven reconfiguration of the brain modular structure over time. Future work could investigate the behavioral relevance of the switching between states of network modularity, e.g., switching rate as a predictive of cognitive ability. Our current framework builds upon a basic SBM which assumes binary networks where edges carry no weights and identical degree distribution of each node. It can be extended to incorporate enhanced variants of SBM, such as the degree-corrected SBM \cite{Karrer2011} to allow for degree heterogeneity within communities, and weighted SBM \cite{Aicher2014} to handle weighted connectivity networks. Another possible extension is to allow both the the connectivity parameters and community memberships to vary over time as in \cite{Xu2014} to capture potential dissolution and formation of new communities. To deal with the arising label switching issue across different time steps \cite{Matias2017}, the proposed multilayer $\text{Q}_{\max}$ can be used to detect the time-varying community partitions by varying the inter-layer coupling parameters, while preserving a consistent mapping of community labels across time.

\vspace{-0.05in}
\bibliographystyle{IEEEbib}
\bibliography{SBM-Ref-IEEE}

\end{document}


\date{}
\title{Supplementary Material: Detecting Dynamic Community Structure in Functional Brain Networks Across Individuals: A Multilayer Approach}

\author{
Chee-Ming Ting\footnote{C-M Ting is with the School of Information Technology, Monash University Malaysia, 47500 Subang Jaya, Malaysia, and also the Biostatistics Group, King Abdullah University of Science \& Technology, Thuwal 23955, Saudi Arabia (e-mail: ting.cheeming@monash.edu).}, S. Balqis Samdin\footnote{S. B. Samdin is with the School of Electrical and Computer Engineering, Xiamen University Malaysia, 43900 Sepang, Malaysia, and also the Biostatistics Group, King Abdullah University of Science \& Technology, Thuwal 23955, Saudi Arabia.}, Meini Tang and Hernando Ombao\footnote{M. Tang and H. Ombao are with the Biostatistics Group, King Abdullah University of Science \& Technology, Thuwal 23955, Saudi Arabia.}
}

\maketitle

This appendix contains additional material to accompany our paper ``Detecting Dynamic Community Structure in Functional Brain Networks Across Individuals: A Multilayer Approach". In Section 1, we provide the details of the experimental design, acquisition and preprocessing of the fMRI data being used. Section 2 presents supplementary results on subject-specific and multi-subject community detection, and analysis of motor task fMRI data. Section 3 presents a series of validation experiments to clarify the choice of important parameters in the estimation. Section 4 and 5 discuss the convergence behavior and computational aspect of the proposed method.

\section{Data Acquisition \& Pre-processing}

We applied the proposed MSS-SBM method to task-related fMRI data of 400 subjects for the language tasks and 450 subjects for the motor tasks from the Human Connectome Project. The data were acquired with 3T Siemens Skyra with TR = 720 ms, TE = 33.1 ms, flip angle = 52\degree, BW =2290 Hz/Px, in-plane FOV = 208 x 180 mm, 72 slices, 2.0 mm isotropic voxels. Refer to \cite{Barch2013} for details of this dataset. 

\textbf{Language data}: The experimental paradigm is block-design alternating between sentence judgments (story) and arithmetic (math) tasks \cite{Binder2011}. Subjects were given four sets of story-math tasks with average duration of approximately 30s for each task. In the story task, subjects were given a short audio with 5-9 sentences followed by a question regarding topic of the story which required the subjects to choose an answer from two given selections. In the math task, the subjects were given a few basic arithmetic questions such as additional and subtraction and followed by answer selection for each question. Each run has $T=316$ time scans and 8 blocks. The onset of task cues can vary across different subjects.

\textbf{Motor data}: The task involves execution of five motor movements: left-hand, left-foot, right-hand, right-foot and tongue movement. Each movement was repeated twice with duration of 12s with cue of 3s prior to the executions. Each dataset has a total of $T=284$ scans, 10 task blocks and 3 fixation blocks per run. Unlike the language tasks, the onset timing of each task was identical for every subject.

Both datasets were minimally preprocessed with structural and functional HCP pre-processing pipelines version of 3.13.2 \cite{Glasser2013}. We used the automated anatomical labeling (AAL) for parcellation of the whole brain into 90 anatomical regions of interest (ROIs) and computed an averaged fMRI time series for each ROI by averaging over voxels.

\section{Supplementary Results}

\subsection{Consensus Community Structure in Single-Subject Analysis}

\begin{figure}[!t]
\centering
\includegraphics[width=0.6\linewidth,keepaspectratio]{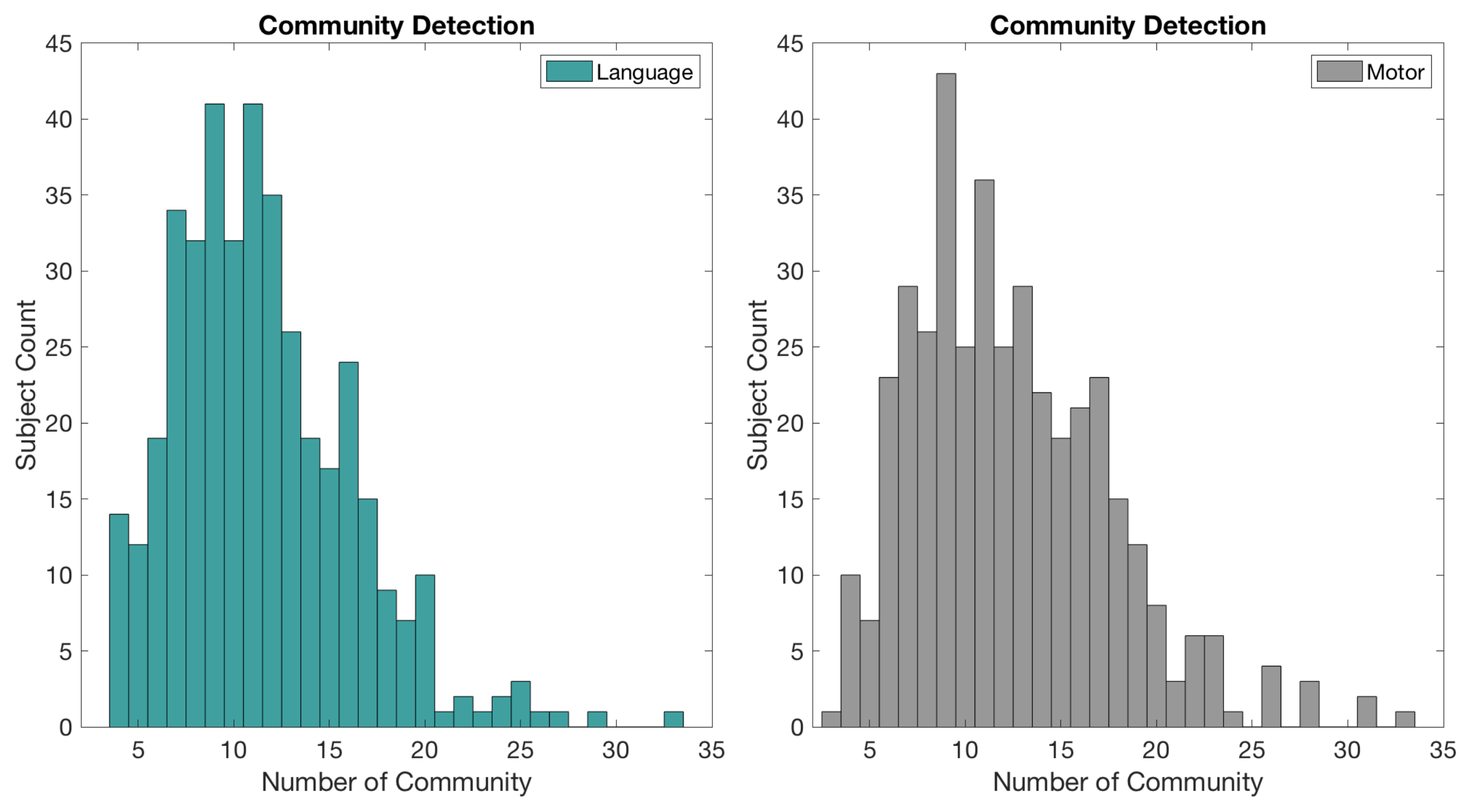}
\vspace{-0.1in}
\caption{Distributions of number of detected brain communities for language (green) and motor (grey) tasks over individual subjects.} 
	\label{Community_detected}
	\vspace{-0.1in}
\end{figure}

\begin{figure*}[!t]
\centering
\includegraphics[width=0.8\linewidth,keepaspectratio]{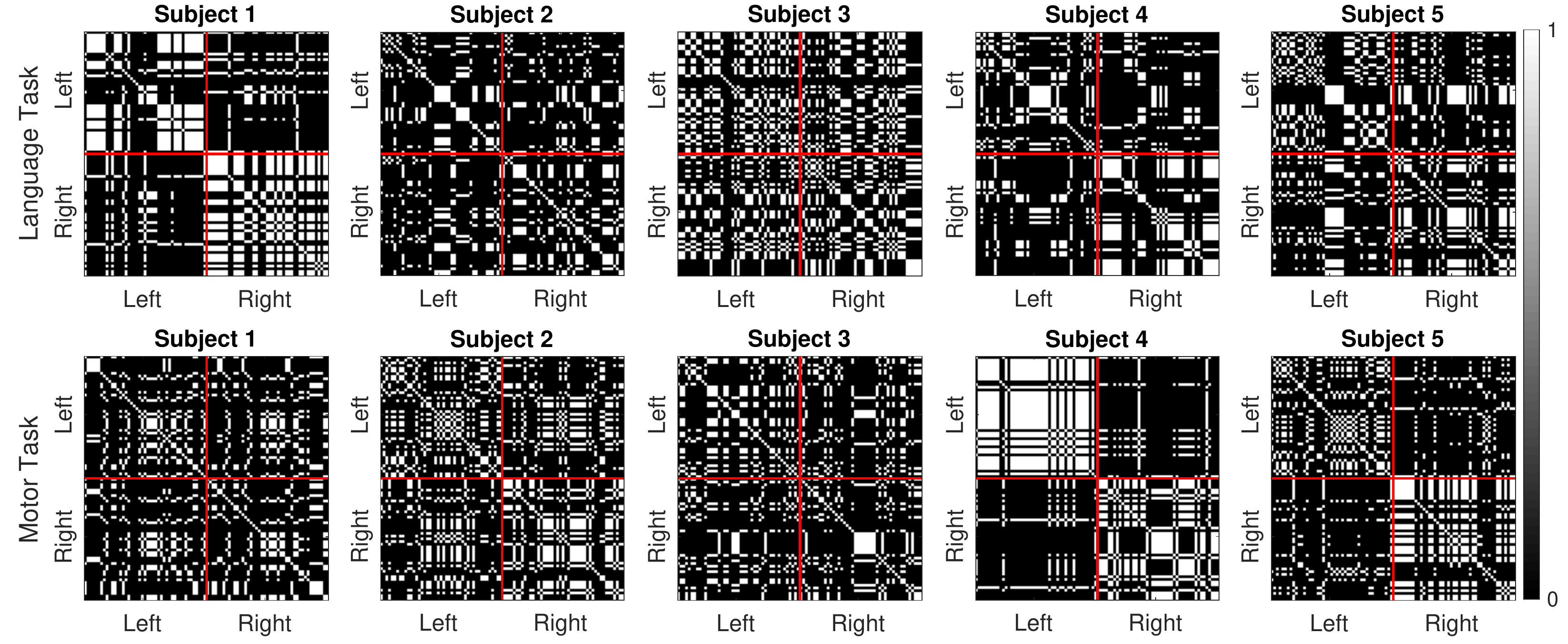} 
\vspace{-0.1in}
	\caption{Community association matrices for 5 subjects for language task (Top) and motor task (Bottom). The ROI-wise entries in the matrices are arranged according to left and right brain hemisphere.}  \label{AsscociateMat}
\end{figure*}

\begin{figure}[!t]
\centering
\hspace{-0.45 cm}
	\begin{minipage}[t]{0.5\linewidth}
		\centering
		\subfloat[Language tasks]{\includegraphics[width=0.7\linewidth,keepaspectratio]{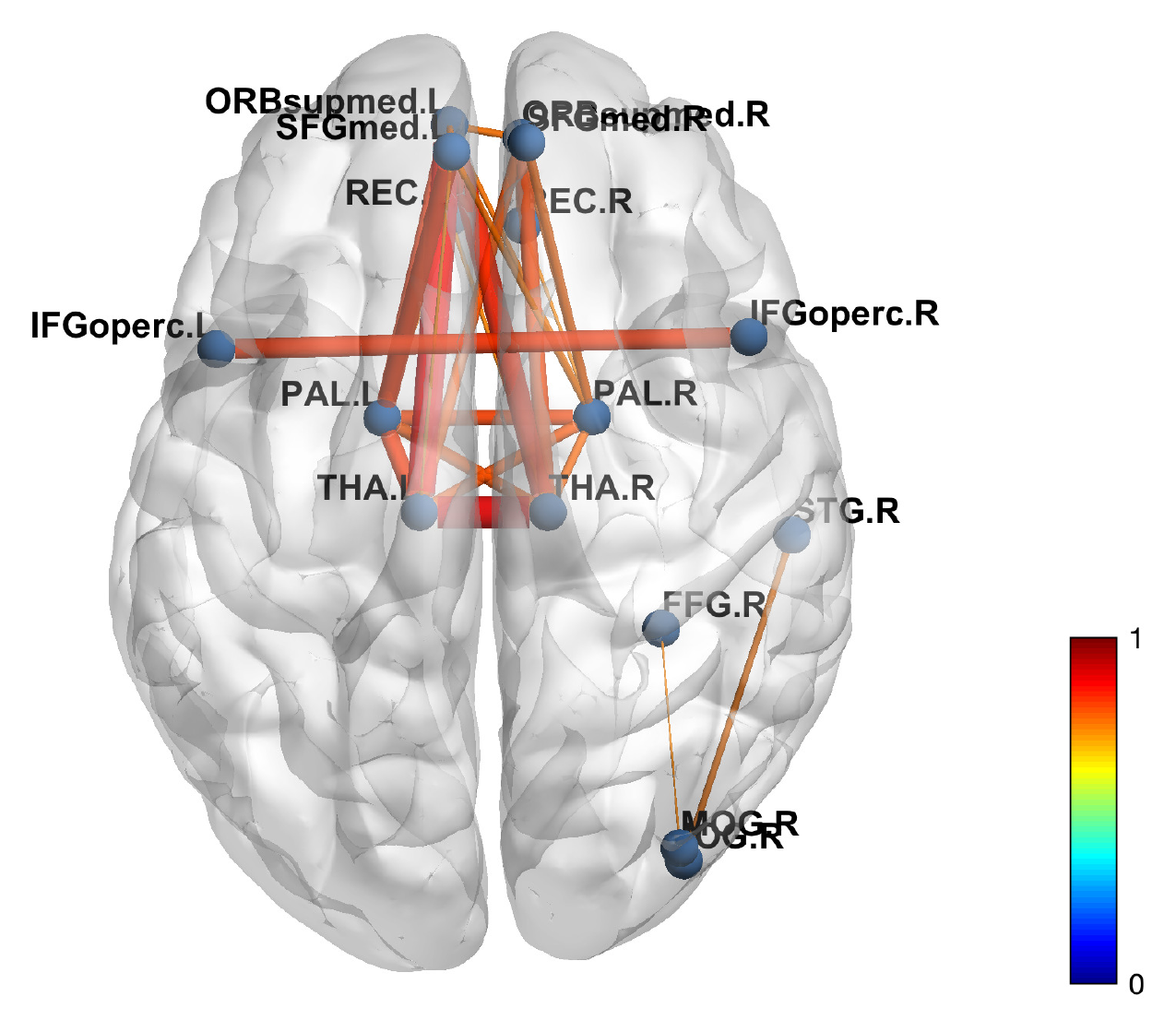}}
	\end{minipage}
	\hspace{-0.7in}
	\begin{minipage}[t]{0.5\linewidth}
		\centering
		\subfloat[Motor tasks]{\includegraphics[width=0.7\linewidth,keepaspectratio]{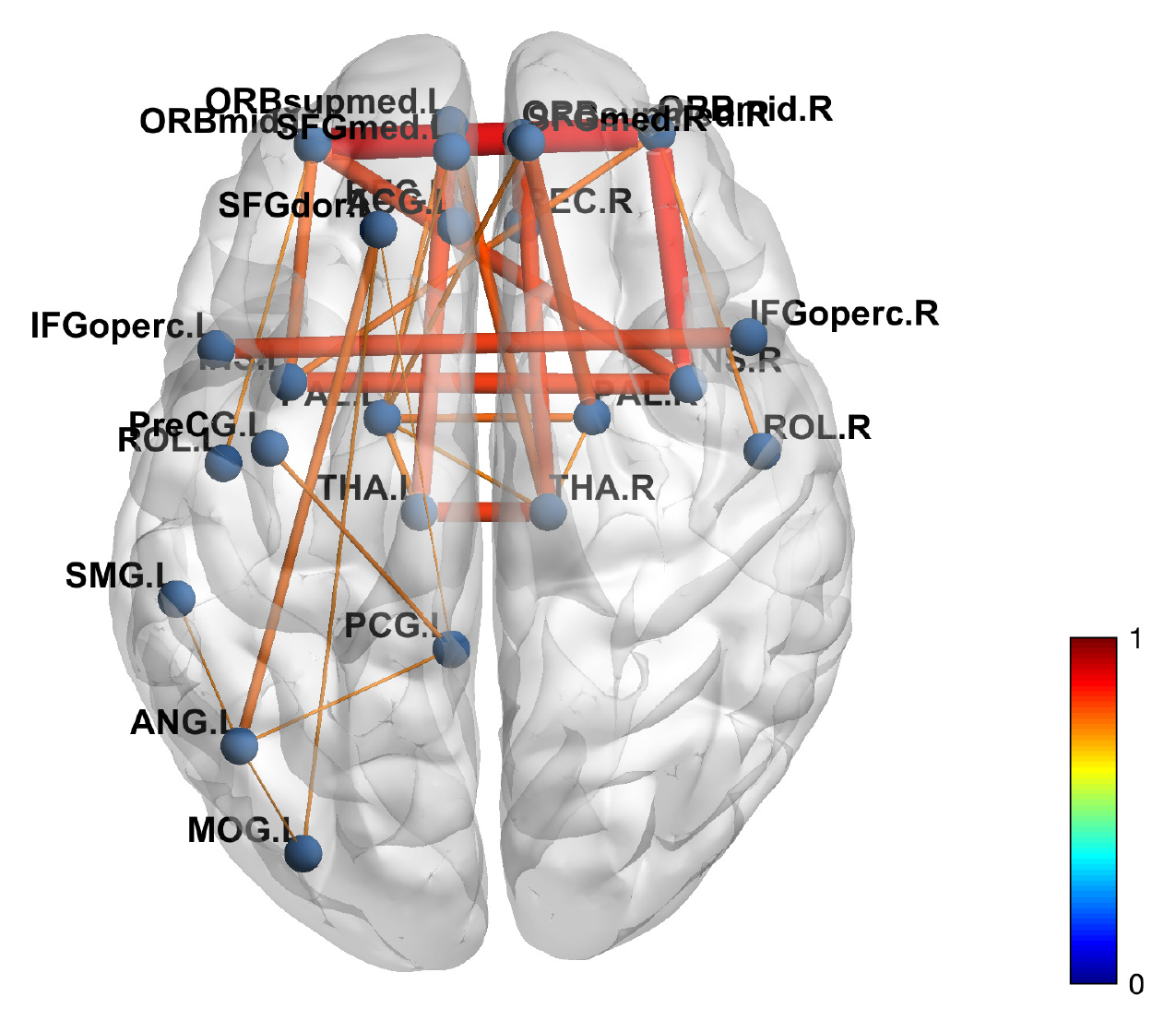}}
	\end{minipage}
\caption{Top 1\% pairs of connected ROIs that are most frequently assigned to the same community over all subjects.}
\label{mapAssociate}
\end{figure}

This section describes a method to examine consensus community partition across subjects each with a different number of communities produced by subject-specific community detection methods. Fig.~\ref{Community_detected} shows distributions of detected number of communities by Louvain algorithm over individual subjects for the language and motor tasks. The results suggest the variability in number of communities across different subjects. To obtain a consistent mapping of community partition across subjects from single-subject community detection by the Louvain algorithm, we compute a $N \times N$ community association matrix $\mathbf{P}_r = [P_{rij}]$ for each subject $r$, where $P_{rij}$ is equal to 1 if nodes $i$ and $j$ are assigned to the same community. Fig.~\ref{AsscociateMat} shows the association matrices for five randomly selected subjects from both datasets. We can investigate the consensus community partition by counting occurrence over subjects that a pair of nodes belong to same community, i.e., $\mathbf{P} = \sum_{r=1}^R \mathbf{P}_r$. Fig.~\ref{mapAssociate} shows the top 1\% pairs of nodes that are most consistently assigned to the same community. We can see community structures for both language and motor tasks involve the frontal and thalamus areas. The motor tasks engage additional regions related to motor control and execution of voluntary movement such as precentral gyrus and rolandic operculum.

\subsection{Identified Community Membership in Multi-Subject Analysis}

Table.~\ref{Table:MODULE-ROI-LANG} and Table.~\ref{Table:MODULE-ROI-MOTOR} list the ROI members of four largest group-wise communities identified based on the multilayer modularity maximization from language and motor fMRI functional networks, respectively.

\begin{table}[!ht]
\caption{Group-wise community membership of brain ROIs for four largest communities identified from language task fMRI over all subjects based on multilayer modularity maximization. The ROIs were parcellated according to Anatomical Automatic Labeling (AAL) atlas.}
\resizebox{0.8\textwidth}{!}{\begin{minipage}{\textwidth}
\begin{center}
\begin{tabular}{lcclcc}
\hline\hline
Membership&ROI Label& &Membership&ROI Label \\
\hline\hline
\multicolumn{2}{l}{\bfseries{Community \#2}} & & L Heschl gyrus & HES.L  \\
R Precentral gyrus & PreCG.R & & L Superior frontal gyrus (dorsal) & SFGdor.L \\
L Rolandic operculum & ROL.L & & R Orbitofrontal cortex (superior) & ORBsup.R \\
R Rolandic operculum & ROL.R & & L Inferior frontal gyrus (opercular) & IFGoperc.L \\
R Supplementary motor area & SMA.R & & L Inferior frontal gyrus (triangular) & IFGtriang.L \\
R Postcentral gyrus & PoCG.R & & L Orbitofrontal cortex (inferior) & ORBinf.L \\
R Superior parietal gyrus & SPG.R & & R Insula & INS.R \\
R Supramarginal gyrus & SMG.R & & L Olfactory & OLF.L \\
R Paracentral lobule & PCL.R & & R Anterior cingulate gyrus & ACG.R \\
R Heschl gyrus & HES.R & & L Middle cingulate gyrus & DCG.L \\
R Superior frontal gyrus (dorsal) & SFGdor.R & & L Posterior cingulate gyrus & PCG.L \\
L Orbitofrontal cortex (superior) & ORBsup.L & & L Hippocampus & HIP.L \\
L Orbitofrontal cortex (middle) & ORBmid.L & & L Parahippocampal gyrus & PHG.L \\
R Orbitofrontal cortex (middle) & ORBmid.R & & L Precuneus & PCUN.L \\
R Inferior frontal gyrus (opercular) & IFGoperc.R & & L Temporal pole (middle) & TPOmid.L \\
R Inferior frontal gyrus (triangular) & IFGtriang.R & & R Temporal pole (middle) & TPOmid.R \\
R Orbitofrontal cortex (inferior) & ORBinf.R & & L Inferior temporal gyrus & ITG.L \\
L Inferior parietal lobule & IPL.L & & L Calcarine cortex & CAL.L \\
R Amygdala & AMYG.R & & L Lingual gyrus & LING.L \\
R Caudate & CAU.R & & L Superior Occipital gyrus & SOG.L \\
L Superior frontal gyrus (medial) & SFGmed.L & & L Middle occipital gyrus & MOG.L \\
R Superior frontal gyrus (medial) & SFGmed.R & & L Inferior occipital gyrus & IOG.L \\
R Orbitofrontal cortex (superior-medial) & ORBsupmed.R & & L Fusiform gyrus & FFG.L \\
R Rectus gyrus & REC.R & & L Superior temporal gyrus & STG.L \\
R Middle cingulate gyrus & DCG.R & & L Temporal pole (superior) & TPOsup.L \\
\cline{3-5}
R Posterior cingulate gyrus & PCG.R & &  \multicolumn{2}{l}{\bfseries{Community \#5}}  \\
R Parahippocampal gyrus & PHG.R & & L Insula & INS.L \\
R Angular gyrus & ANG.R & & R Inferior parietal lobule & IPL.R \\
R Precuneus & PCUN.R & & R Olfactory & OLF.R \\
R Inferior temporal gyrus & ITG.R & & L Pallidum & PAL.L \\
R Calcarine cortex & CAL.R & & R Pallidum & PAL.R \\
R Lingual gyrus & LING.R & & L Thalamus & THA.L \\
R Superior occipital gyrus & SOG.R & & R Thalamus & THA.R \\
R Middle occipital gyrus & MOG.R & & L Orbitofrontal cortex (superior-medial) & ORBsupmed.L \\
R Fusiform gyrus & FFG.R & & L Rectus gyrus & REC.L \\
R Temporal pole (superior) & TPOsup.R & & L Anterior cingulate gyrus & ACG.L \\
\cline{1-2}
\multicolumn{2}{l}{\bfseries{Community \#1}} & & R Hippocampus & HIP.R  \\
L Precentral gyrus & PreCG.L & & L Angular gyrus & ANG.L \\
L Supplementary motor area & SMA.L & & L Middle temporal gyrus & MTG.L \\
L Postcentral gyrus & PoCG.L & & R Middle temporal gyrus & MTG.R \\
\cline{3-5}
L Superior parietal gyrus & SPG.L & &  \multicolumn{2}{l}{\bfseries{Community \#12}}  \\
L Supramarginal gyrus & SMG.L & & R Inferior occipital gyrus & IOG.R \\
L Paracentral lobule & PCL.L & & R Superior temporal gyrus & STG.R \\
\hline\hline
\end{tabular}
\end{center}
\label{Table:MODULE-ROI-LANG}
\end{minipage} }
\end{table}

\begin{table}[!ht]
\caption{Group-wise community membership of brain ROIs for four largest communities identified from motor task fMRI over all subjects based on multilayer modularity maximization. The ROIs were parcellated according to Anatomical Automatic Labeling (AAL) atlas.}
\resizebox{0.85\textwidth}{!}{\begin{minipage}{\textwidth}
\begin{center}
\begin{tabular}{lcclcc}
\hline\hline
Membership&ROI Label& &Membership&ROI Label \\
\hline\hline
\multicolumn{2}{l}{\bfseries{Community \#1}} & & L Precuneus & PCUN.L \\
 L Precentral gyrus & PreCG.L & & R Precuneus & PCUN.R \\
 R Precentral gyrus & PreCG.R & & R Superior occipital gyrus & SOG.R \\
 L Supplementary motor area & SMA.L & & R Middle occipital gyrus & MOG.R \\
 L Superior parietal gyrus & SPG.L & & L Inferior occipital gyrus & IOG.L \\
 L Orbitofrontal cortex (superior) & ORBsup.L & & R Inferior occipital gyrus & IOG.R \\
 R Orbitofrontal cortex (superior) & ORBsup.R & & L Fusiform gyrus & FFG.L \\
 L Olfactory & OLF.L & & R Fusiform gyrus & FFG.R \\
 R Olfactory & OLF.R & & R Superior temporal gyrus & STG.R \\
\cline{3-5}
 R Pallidum & PAL.R & &  \multicolumn{2}{l}{\bfseries{Community \#5}}   \\
 L Thalamus & THA.L & & L Superior frontal gyrus (dorsal) & SFGdor.L \\
 R Thalamus & THA.R & & R Superior frontal gyrus (dorsal) & SFGdor.R \\
 L Orbitofrontal cortex (superior-medial) & ORBsupmed.L & & L Inferior parietal lobule & IPL.L \\
 R Orbitofrontal cortex (superior-medial) & ORBsupmed.R & & R Inferior parietal lobule & IPL.R \\
 L Rectus gyrus & REC.L & & L Pallidum & PAL.L \\
 L Anterior cingulate gyrus & ACG.L & & L Superior frontal gyrus (medial) & SFGmed.L \\
 R Anterior cingulate gyrus & ACG.R & & R Superior frontal gyrus (medial) & SFGmed.R \\
 L Middle cingulate gyrus & DCG.L & & R Middle cingulate gyrus & DCG.R \\
 R Posterior cingulate gyrus & PCG.R & & L Posterior cingulate gyrus & PCG.L \\
 L Hippocampus & HIP.L & & L Inferior temporal gyrus & ITG.L \\
 R Hippocampus & HIP.R & & R Inferior temporal gyrus & ITG.R \\
 L Middle temporal gyrus & MTG.L & & L Calcarine cortex & CAL.L \\
 R Middle temporal gyrus & MTG.R & & R Calcarine cortex & CAL.R \\
 L Temporal pole (middle) & TPOmid.L & & R Lingual gyrus & LING.R \\
 R Temporal pole (middle) & TPOmid.R & & L Superior occipital gyrus & SOG.L \\
 L Lingual gyrus & LING.L & & L Middle occipital gyrus & MOG.L \\
 R Temporal pole (superior) & TPOsup.R & & L Temporal pole (superior) & TPOsup.L \\
\cline{1-2} \cline{3-5}
\multicolumn{2}{l}{\bfseries{Community \#2}} & & \multicolumn{2}{l}{\bfseries{Community \#4}}  \\
 L Rolandic operculum & ROL.L & & L Postcentral gyrus & PoCG.L \\
 R Rolandic operculum & ROL.R & & L Supramarginal gyrus & SMG.L \\
 R Postcentral gyrus & PoCG.R & & R Supramarginal gyrus & SMG.R \\
 R Superior parietal gyrus & SPG.R & & L Paracentral lobule & PCL.L \\
 L Heschl gyrus & HES.L & & R Paracentral lobule & PCL.R \\
 R Heschl gyrus & HES.R & & L Orbitofrontal cortex (inferior) & ORBinf.L \\
 L Orbitofrontal cortex (middle) & ORBmid.L & & R Orbitofrontal cortex (inferior) & ORBinf.R \\
 R Orbitofrontal cortex (middle) & ORBmid.R & & R Rectus gyrus & REC.R \\
 L Inferior frontal gyrus (opercular) & IFGoperc.L & & L Parahippocampal gyrus & PHG.L \\
 R Inferior frontal gyrus (opercular) & IFGoperc.R & & R Parahippocampal gyrus & PHG.R \\
 L Insula & INS.L & & L Angular gyrus & L ANG.L \\
 R Insula & INS.R & & R Angular gyrus & R ANG.R \\

\hline\hline
\end{tabular}
\end{center}
\label{Table:MODULE-ROI-MOTOR}
\end{minipage} }
\end{table}

\subsection{Results for Motor fMRI Data}

In the main text, we presented the results of detecting dynamic community structure in the fMRI functional networks using the proposed MSS-SBM for the language task. In this section, we present a similar analysis with the motor task.

\begin{figure}[!t]
\centering
	\includegraphics[width=0.72\linewidth,keepaspectratio]{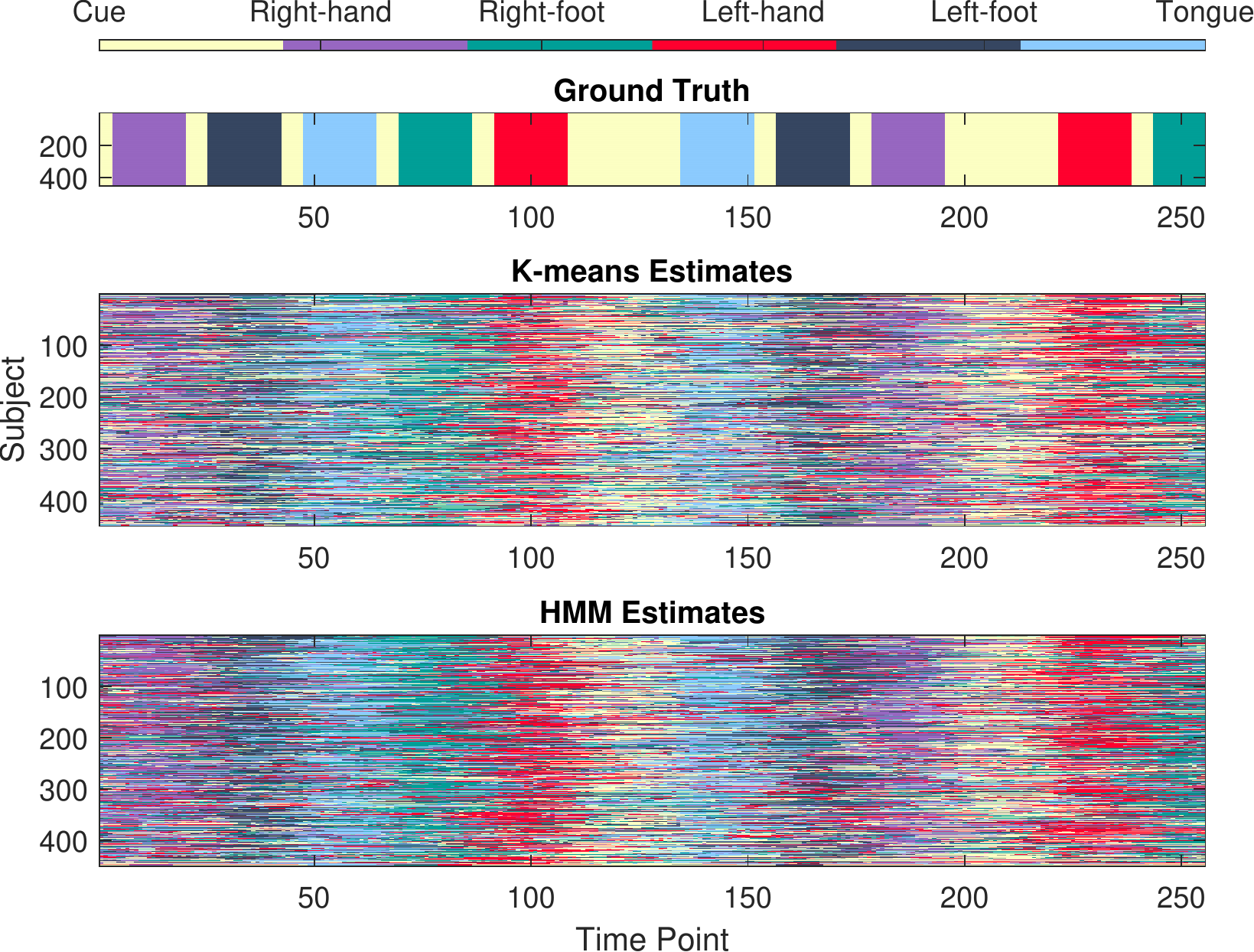}
\caption{Tracking of regime changes in the inter-community connectivity of fMRI functional networks over individual subjects for motor task experiments. (Top) Ground-truth task state time courses from experimental design. (Middle \& Bottom) Dynamic state estimates by K-means clustering and MS-SBM.} \label{MotorTemStateAcc}
\end{figure}

\textbf{Single-Subject Analysis}: For each subject, we fitted MS-SBM with $S = 6$ states on the dynamic functional networks estimated from the motor task fMRI data. The community membership of nodes were obtained by applying the Louvain algorithm on individual subject basis. Fig.~\ref{MotorTemStateAcc} shows the superior performance of the MS-SBM over the K-means clustering in tracking the transitions of the inter-modular connectivity states between the six conditions in the motor task.

\begin{figure}[!t]
\centering
\includegraphics[width=0.6\linewidth,keepaspectratio]{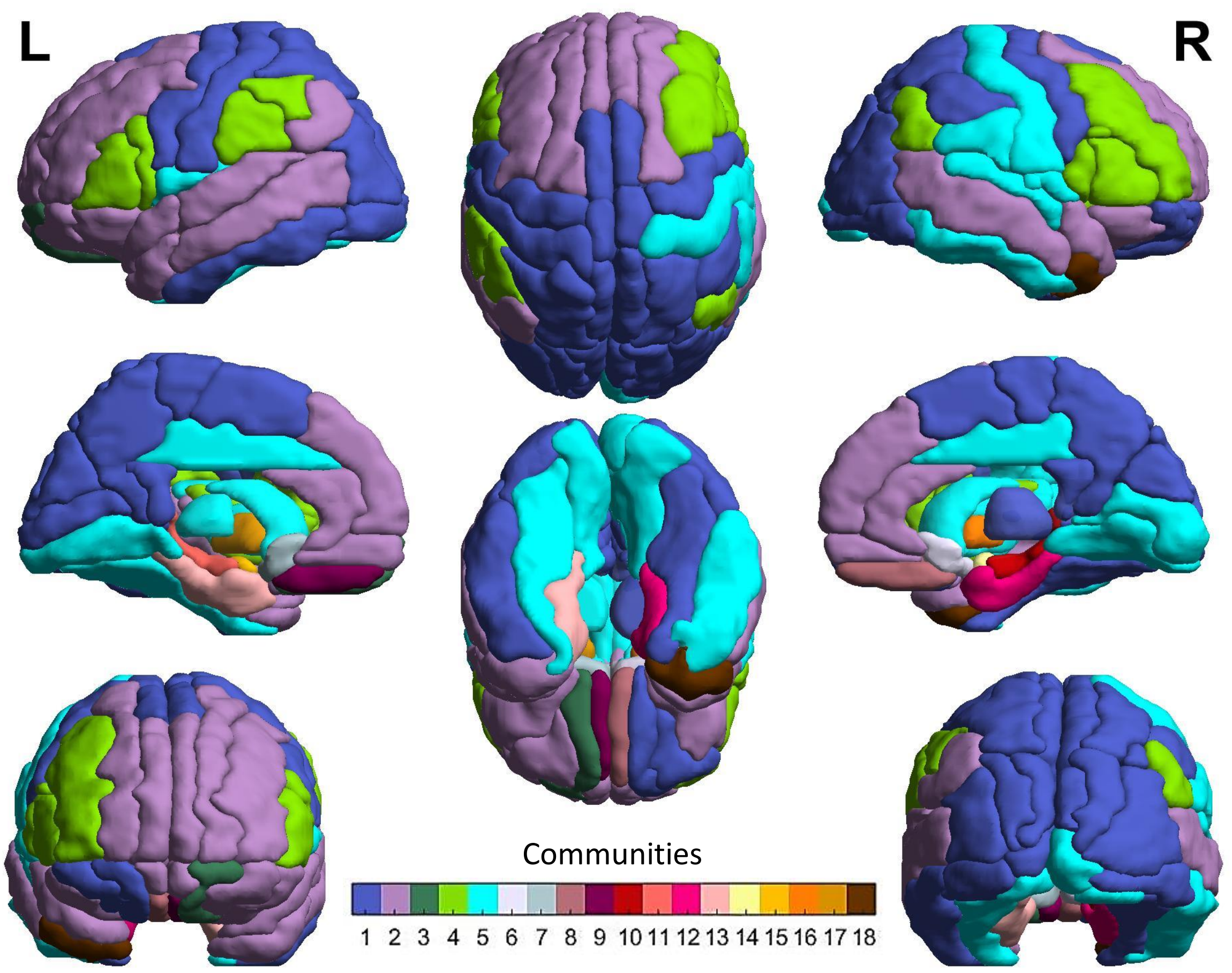} 
	\caption{Topographic representation of group-level brain community partition detected over 450 subjects from the motor task fMRI based on multilayer modularity maximization. The 90 ROIs were color-coded according to their assigned communities.}\label{MotorCommMulti}
	\vspace{0.15in}
\end{figure}

\textbf{Multi-Subject Analysis}: Similar to language task, we applied the proposed MSS-SBM with multilayer $\text{Q}_{\max}$ algorithm for group-level community detection on the motor task data. Fig.~\ref{MotorCommMulti} shows the detected community structure for the motor task, where 18 communities are identified. These include 4 large communities (Community \#1, \#2, \#4 and \#5, see ROI members in Table.~\ref{Table:MODULE-ROI-MOTOR}) and 14 singleton communities. Community \#1 consists of major motor areas including the primary motor cortex (precentral gyrus (PreCG), both left and right) and supplementary motor areas (SMA). It also includes regions of parietal lobe which interacts with other regions such as motor cortex by integrating visual-motor information for movement control \cite{Fogassi2005}. Community \#2 mainly comprises frontal and occipital regions.

\begin{figure}[!t]
		\centering
		\includegraphics[width=1\linewidth,keepaspectratio]{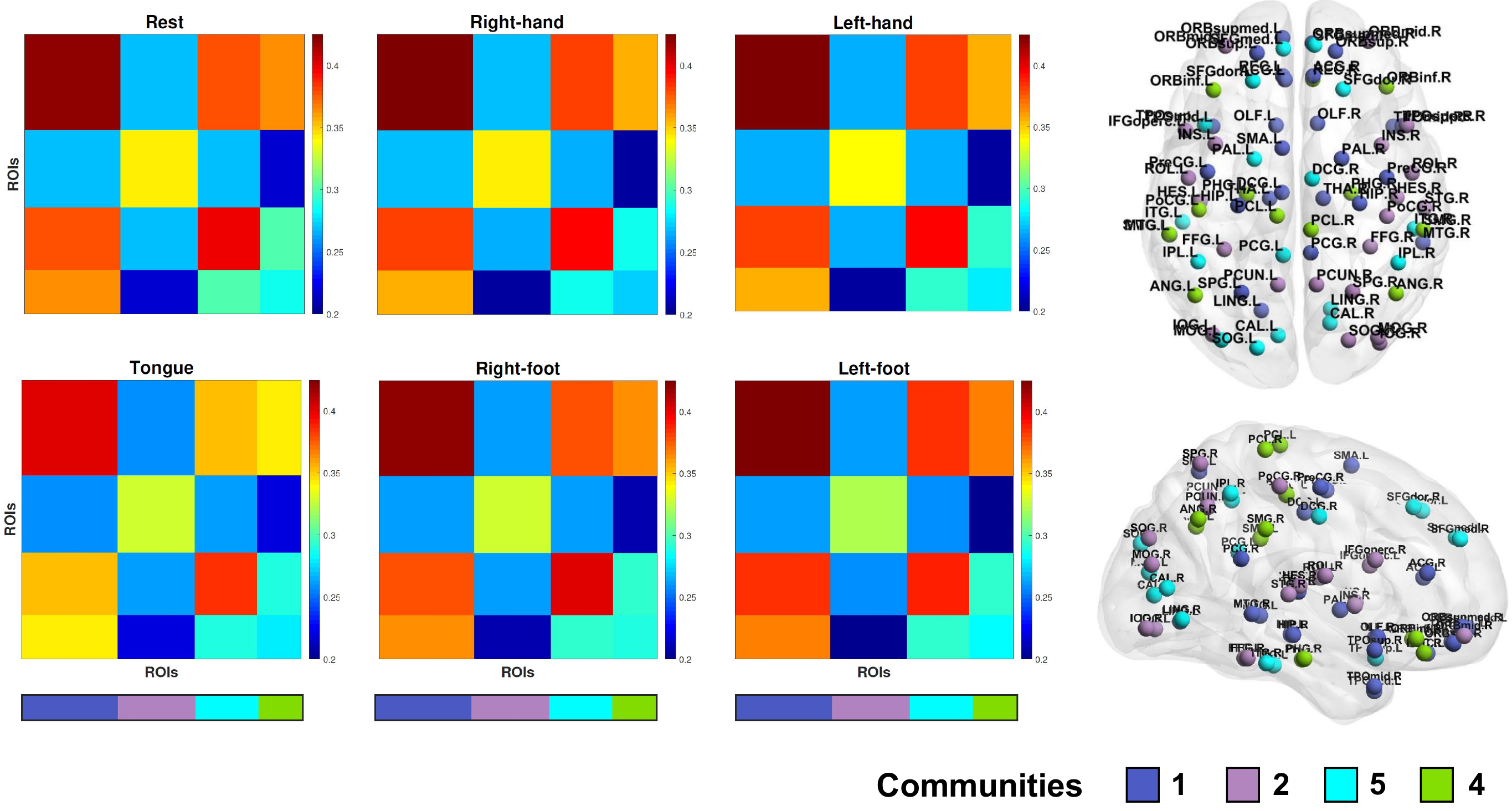}
	\caption{Estimated median modular connection probability matrices between four largest communities for six states in the motor tasks. The spatial distributions of the member ROIs of the communities are depicted in brain renders.}\label{Motor-Densitymat}
\end{figure}

\begin{figure*}[!th]
\centering
\includegraphics[width=0.98\linewidth,keepaspectratio]{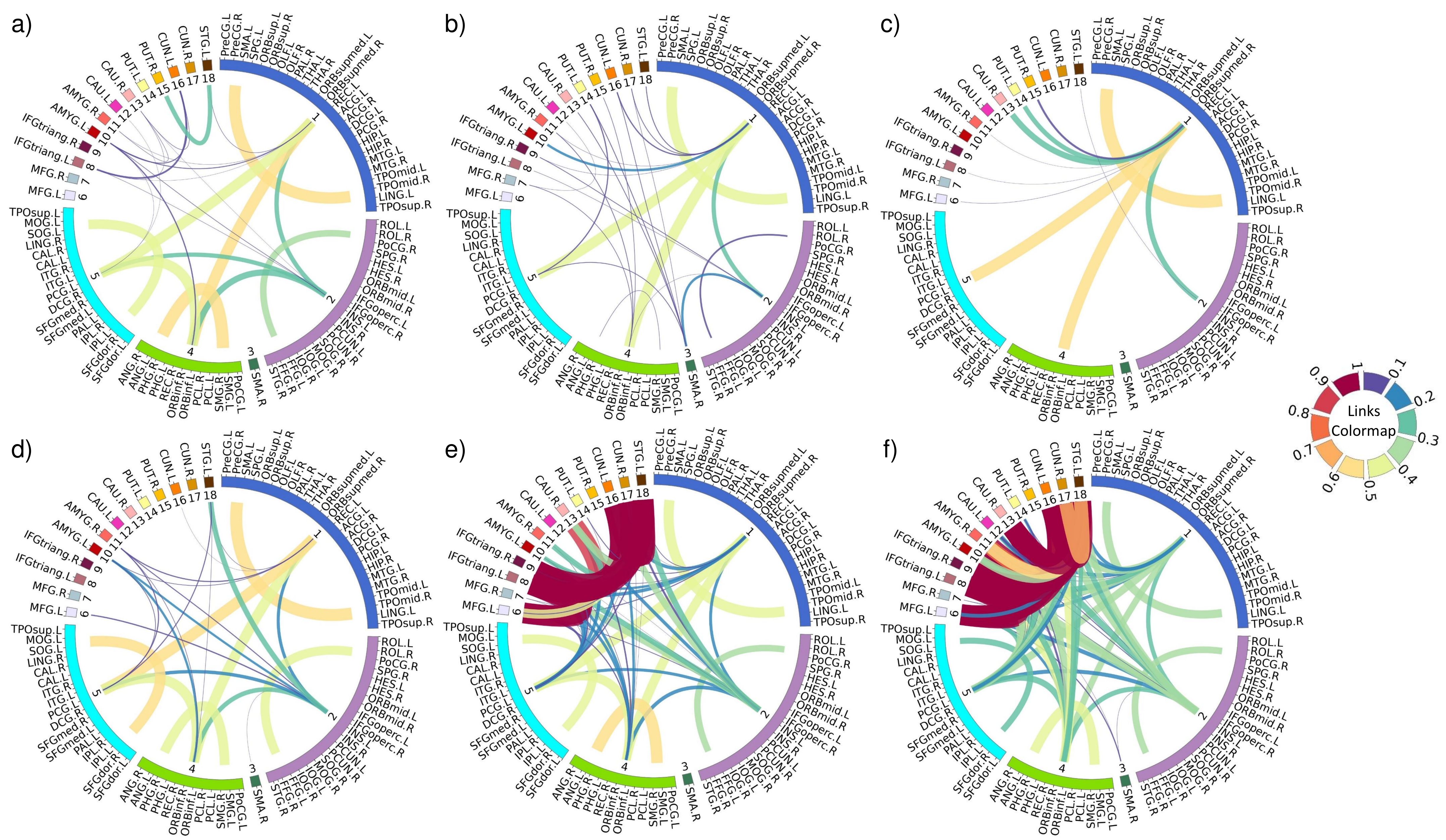} 
	\caption{Within and between-community connectivity for each state of the motor tasks. (a) Rest. (b) Right-hand. (c) Left-hand. (d) Tongue. (e) Right-foot. (f) Left-foot} \label{MotorMap}
	\vspace{0.2in}
\end{figure*}

Fig.~\ref{Motor-Densitymat} shows estimated inter-modular connection probability matrices between four largest communities for the six conditions in the motor task. The non-assortative community motif is evident for all the six states. Community \#1 serves as the core interacting with periphery communities \#4 and \#5. Community \#4 is less assortative with low within-community connection density. The tongue movement has its own unique connectivity pattern. Hand movements exhibit denser connections within community \#2 than foot movements. Fig.~\ref{MotorMap} show interactions between all communities detected for the motor tasks for a subject with the best state alignment with experimental task sequence. We found increase in connectivity within and between communities in the foot movements compared to the hand and tongue movements. Moreover, the left and right movements also exhibit distinct network configurations.

\subsection{Comparison of Different Methods}

\begin{table}[!ht]
\renewcommand{\arraystretch}{1.1}
\caption{Comparison of average (median) Rand indices (RI) and $F1$ scores over subjects of different methods for dynamic modular state identification on language and motor task fMRI.}
\vspace{-0.1 cm}
\label{Table:RI-F1}
\centering
\begin{tabular}{p{2.7cm}p{2.5cm}cccc}
\hline \hline
Community & State & \multicolumn{2}{c}{Language}  & \multicolumn{2}{c}{Motor} \\
\cline{3-6}
Detection & Identification & RI & $F1$ & RI & $F1$ \\
\hline
\multirow{2}{*}{Subject-specific} & K-means & 0.5158 (0.5049) & 0.5023 (0.5014) & 0.6796 (0.6854) & 0.2340 (0.2314)\\
& HMM & 0.5203 (0.5057) & 0.5364 (0.5353) & 0.7105 (0.7164) & 0.2512 (0.2482)\\
\multirow{2}{*}{Multi-subject} & K-means &  0.5204 (0.5085) &  0.5194 (0.5240) & 0.6744 (0.6851) & 0.2594 (0.2570) \\
& HMM & 0.5245 (0.5106) & 0.5492 (0.5546) & 0.7271 (0.7280) & 0.2799 (0.2770) \\
\hline \hline
\end{tabular}
\vspace{0.2in}
\end{table}

\begin{figure}[!t]
\centering
\includegraphics[width=0.6\linewidth,keepaspectratio]{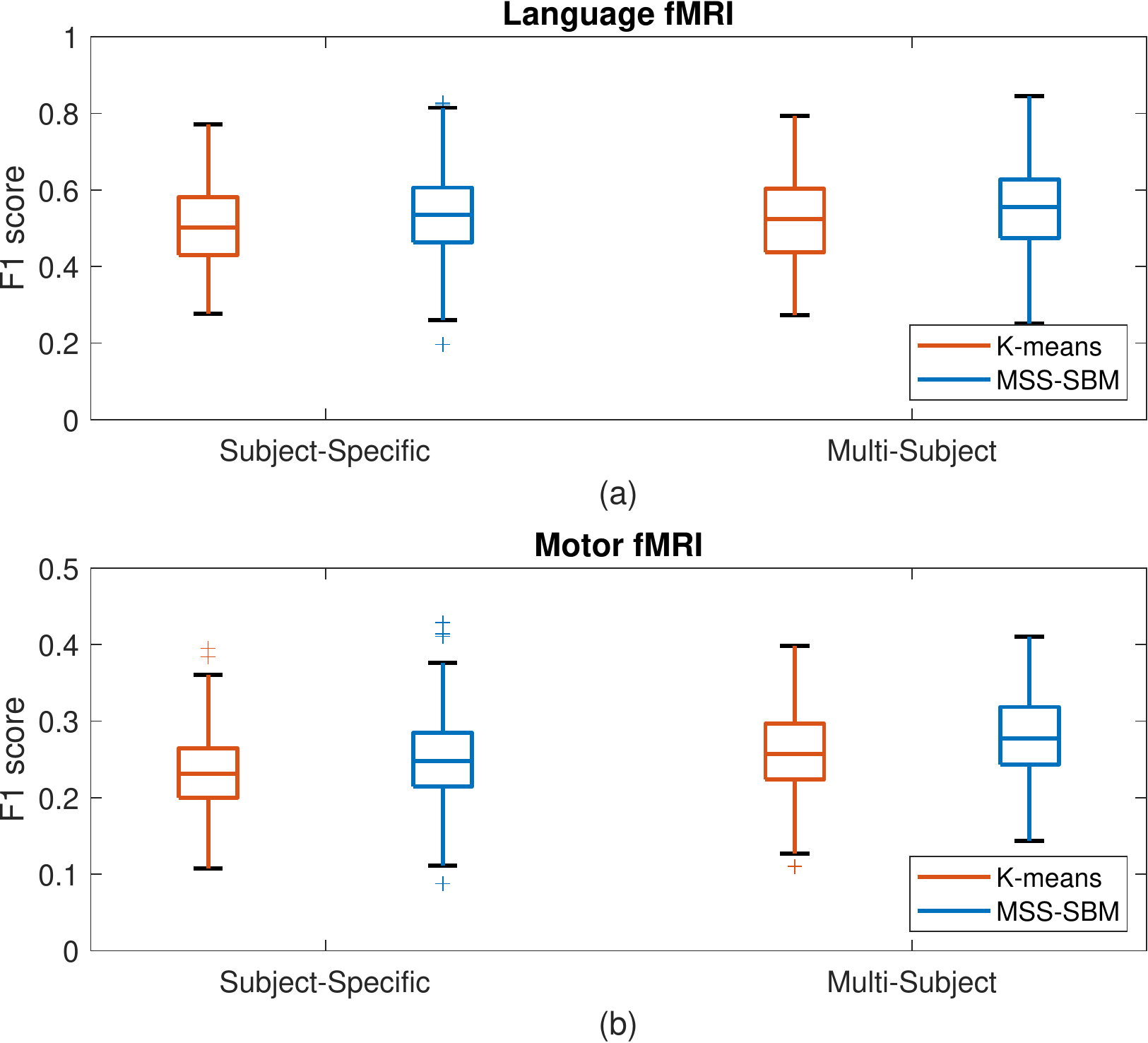} 
\vspace{-0.1in}
	 \caption{Comparison of different methods for tracking modular connectivity state dynamics in language and motor task fMRI, as measured by $F1$ scores of estimated state time courses relative to the experimental ground-truth over all subjects.} \label{F1-score_fmri}
\end{figure}

In the main manuscript, we compared performance of different community detection and temporal clustering algorithms for estimating the modular connectivity state dynamics based on Rand indices (RIs). To further validate the results, we consider an additional performance metric using the $F$-measure \cite{Larsen1999} to compare the estimated state sequence with the ground-truth from experimental paradigm. $F$-measure (or $F1$ score) is a type of cluster evaluation measure based on set overlaps between clusterings by counting pairs of data points which two clusterings have in common (or number of points in the intersection of two clusters). It takes on values in the range $[0,1]$ with value of 1 indicates identical clusterings. Specifically, $F$-measure gives an indication of the accuracy of identified clusters with respect to a ground-truth, calculated as the harmonic mean of precision (fraction of pairs correctly assigned to the same cluster with ground-truth among the identified) and recall (fraction of correct pairs that were identified). Fig.~\ref{F1-score_fmri} plots the $F1$ scores of state assignments of time points for different algorithms. In Table~\ref{Table:RI-F1}, we also report the averaged RIs and $F1$ scores over all subjects. The results confirm the more accurate dynamic state identification of the HMM over the K-means clustering, and multi-subject over the single-subject community detection.

\newpage
\section{Choice of Parameters}

We present a series of additional analyses to clarify the choice of some important parameters which needs to be pre-specified in the MSS-SBM estimation.

\subsection{Edge Density in Network Construction}

In the network construction, we applied the proportional thresholding which defines the threshold $\tau_{\kappa}$ on the correlation matrices in terms of a fixed connection (edge) density $\kappa$ (also called topological cost) of the resulting graphs across individual and datasets. The common practice in brain connectome studies for selecting a proper threshold is to explore network properties as a function of changing threshold or connection density, and identify a range of values (or a single value) over which the resulting network exhibits desirable topological characteristics \cite{Achard2007,Bassett2008,Bullmore2011}. We aim to construct networks with an optimal balance of functional segregation and integration. To this end, we considered the well-known small-world property which combines the presence of functionally specialized (segregated) modules with a robust number of intermodular (integrating) links, and is found ubiquitous in brain networks \cite{Rubinov2010}.

\begin{figure*}[!ht]
\centering
\subfigure[]{\includegraphics[width=0.3\linewidth,keepaspectratio]{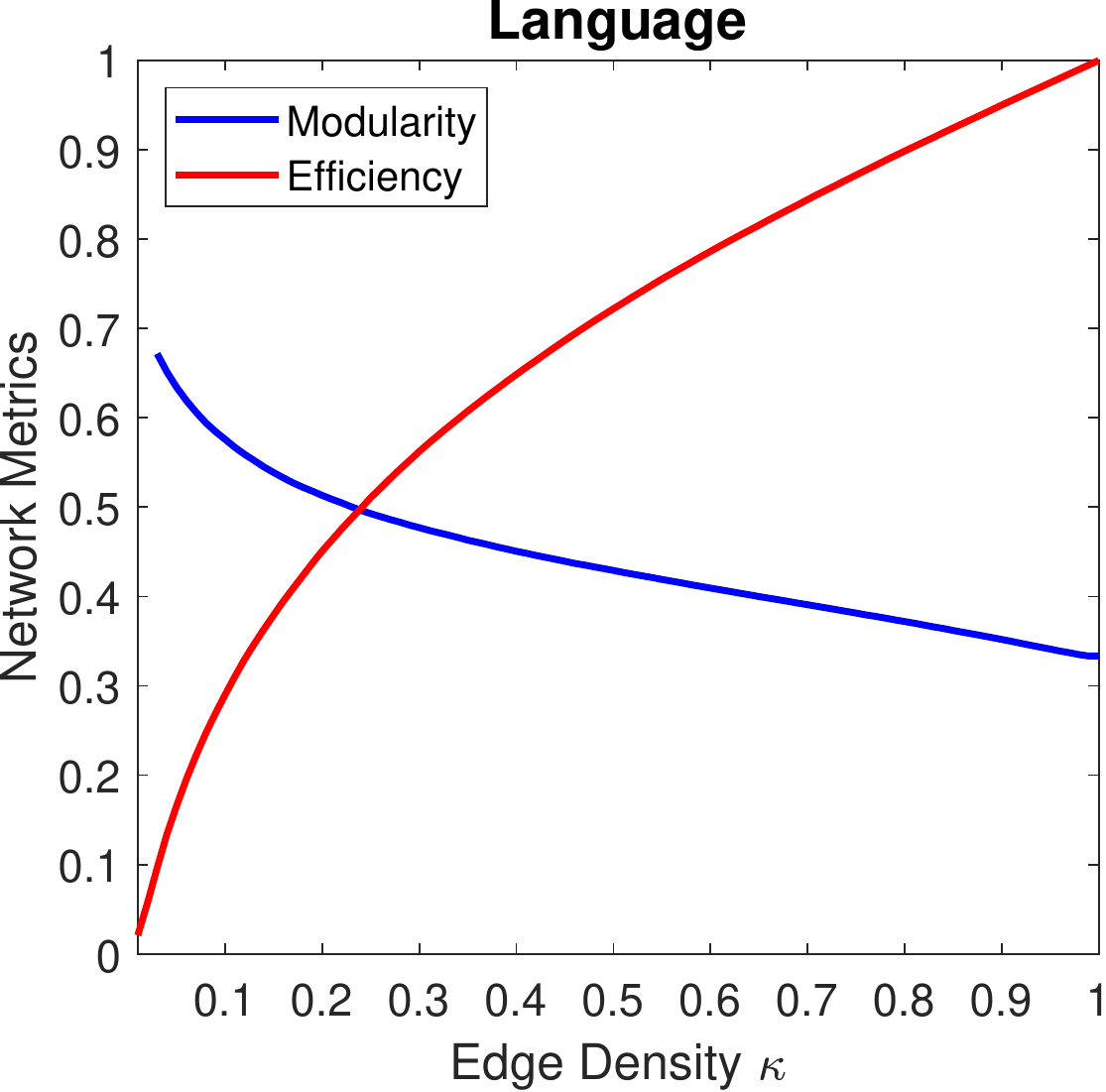}} \hspace{0.4cm}
\subfigure[]{\includegraphics[width=0.302\linewidth,keepaspectratio]{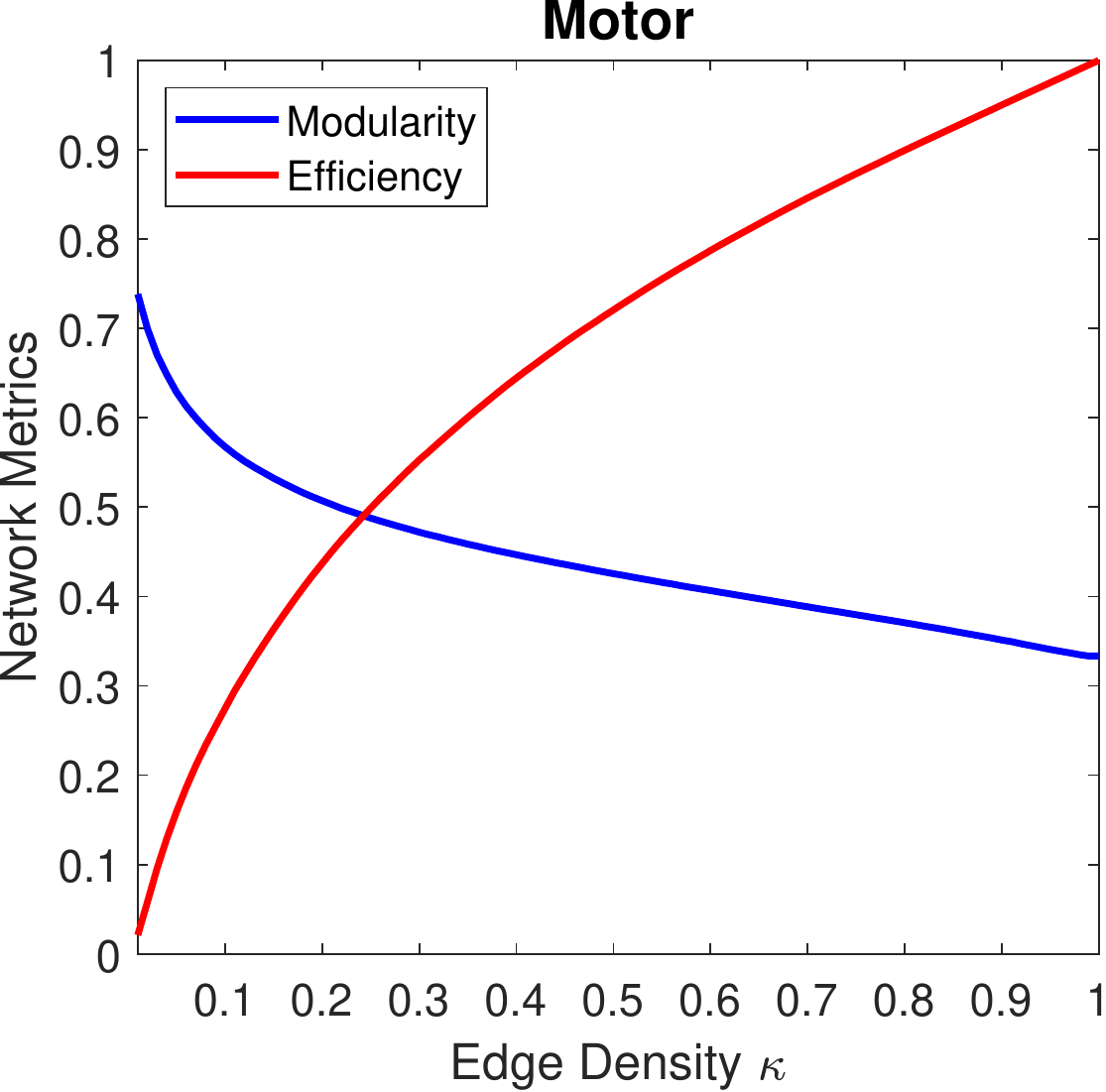}} \vspace{-0.2cm} \\
\subfigure[]{\includegraphics[width=0.3\linewidth,keepaspectratio]{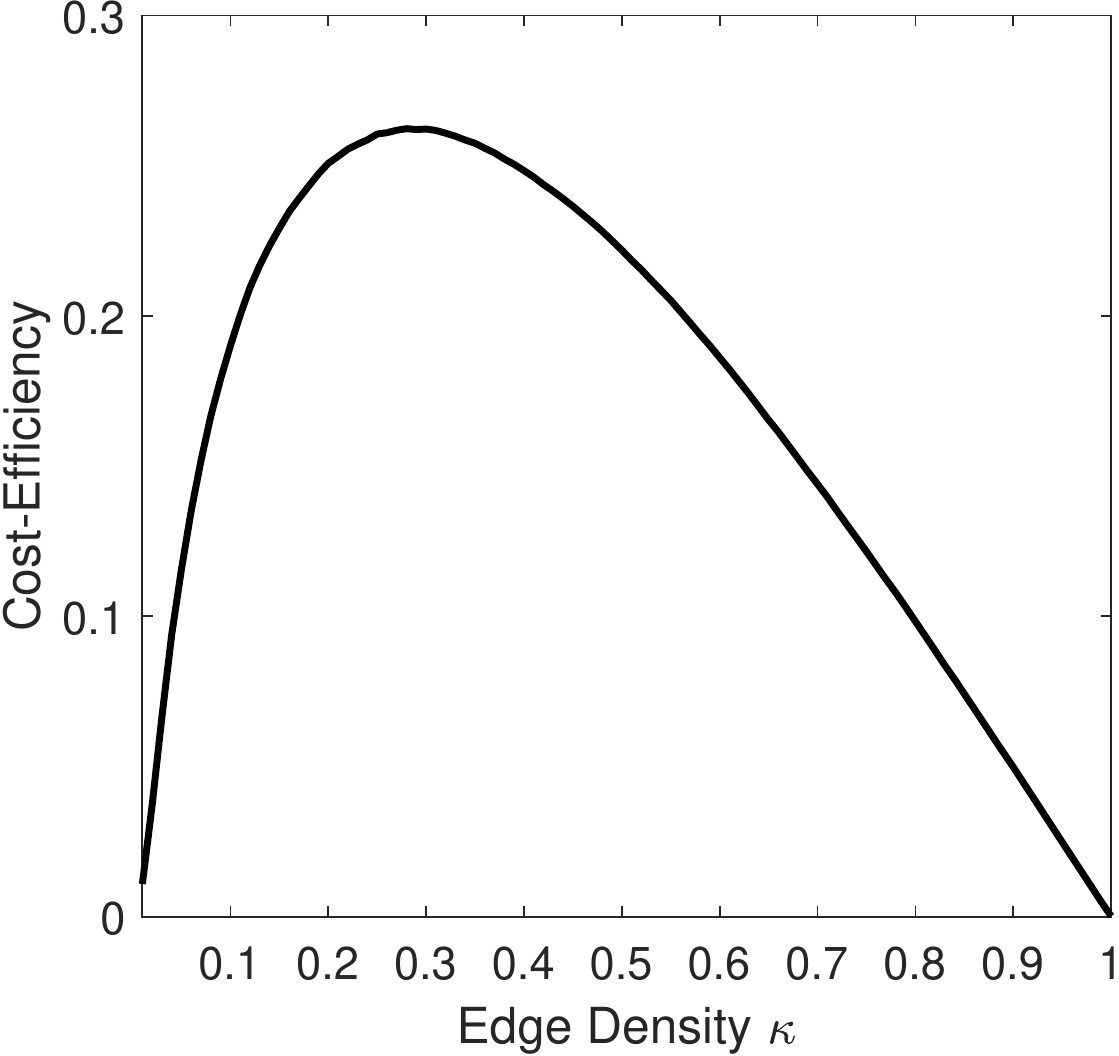}}  \hspace{0.4cm}
\subfigure[]{\includegraphics[width=0.302\linewidth,keepaspectratio]{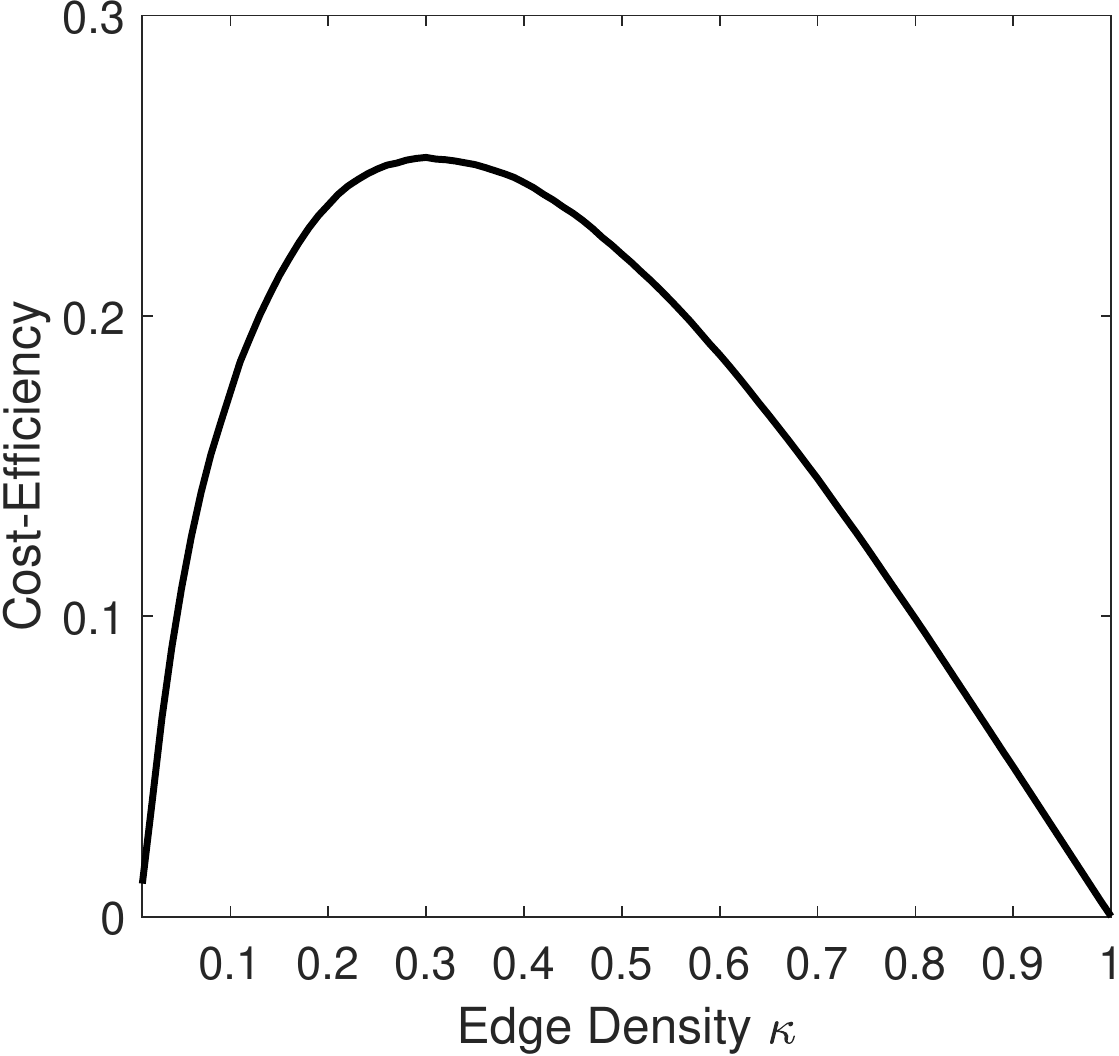}}
\vspace{-0.1in}
\caption{Effect of connection density on topological measures of fMRI functional brain networks for language and motor tasks. (a)-(b) Global efficiency and modularity. (c)-(d) Cost-efficiency. The results are averages over all subjects. Edge density $\kappa = 0.25$ gives the highest cost-efficiency.}
\label{Fig:Eff-kappa}
\end{figure*}

The network \textit{efficiency} can be used as a unified measure to define small-world networks for having both the high global and local efficiency of parallel information transfer \cite{Latora2001}. To investigate the effect of connection density on network efficiency, we varied $\kappa$ from 0.01 to 1 with an increment of 0.01. As expected, the global efficiency increases with the increase of $\kappa$ in the fMRI functional networks for both tasks, as shown in Fig.~\ref{Fig:Eff-kappa}(a)-(b)). This is because more edges facilitate more efficient information flows. However, addition of extra edges along with the increase of network cost $\kappa$, results in relatively modest increments of network efficiency, as evident for $\kappa>0.5$. Based on this observation, \cite{Achard2007} proposed the so-called cost-efficiency to quantify the economical, small-world property in brain functional networks (i.e., optimal in the sense of having high efficiency of information processing at low connection cost). More precisely, the cost-efficiency is defined as the difference between global efficiency and topological cost, which is typically positive and has a positive maximum value at a critical connection density. Using this metric, we can find an optimal value for the threshold
\begin{equation}
\kappa^{*} = \argmax_{\kappa} \{E_{glob}(\kappa) - \kappa \}\label{eq:cost-eff}
\end{equation}
where $E_{glob} = \frac{1}{N}\sum_{i}{\frac{\sum_{j: j\ne i}{(d_{ij})^{-1}}}{N-1}}$ is global efficiency, and $d_{ij}$ is the shortest path length between nodes $i$ and $j$. In Fig.~\ref{Fig:Eff-kappa}(c)-(d)), the plots of cost-efficiency versus $\kappa$ suggest an optimal threshold according to (\ref{eq:cost-eff}) at approximately $\kappa^{*}=0.25$ for both tasks, as indicated by the maximum value of cost-efficiency. The proposed choice of $\kappa=0.25$ is also within the small-word cost regime $0.15 \leq \kappa \leq 0.25$ identified by \cite{Bassett2008} to be appropriate for having fully-connected small-world brain networks, according to the classical metric of small-worldness \cite{Humphries2006,Watts1998}. In contrast to efficiency, the modularity index of the networks ($Q \in [-0.5,1]$ linearly remapped to match the $E_{glob} \in [0,1]$) drops with $\kappa$ (Fig.~\ref{Fig:Eff-kappa}(a)-(b))), indicating reduced modular segregation due to denser inter-modular connections. An optimal tradeoff was identified at which $Q = E_{glob}$, further supporting the proposed choice of $\kappa=0.25$ in order to provide a balance between network segregation and integration.

\begin{figure*}[!t]
\centering
\subfigure[]{\includegraphics[width=0.6\textwidth,keepaspectratio]{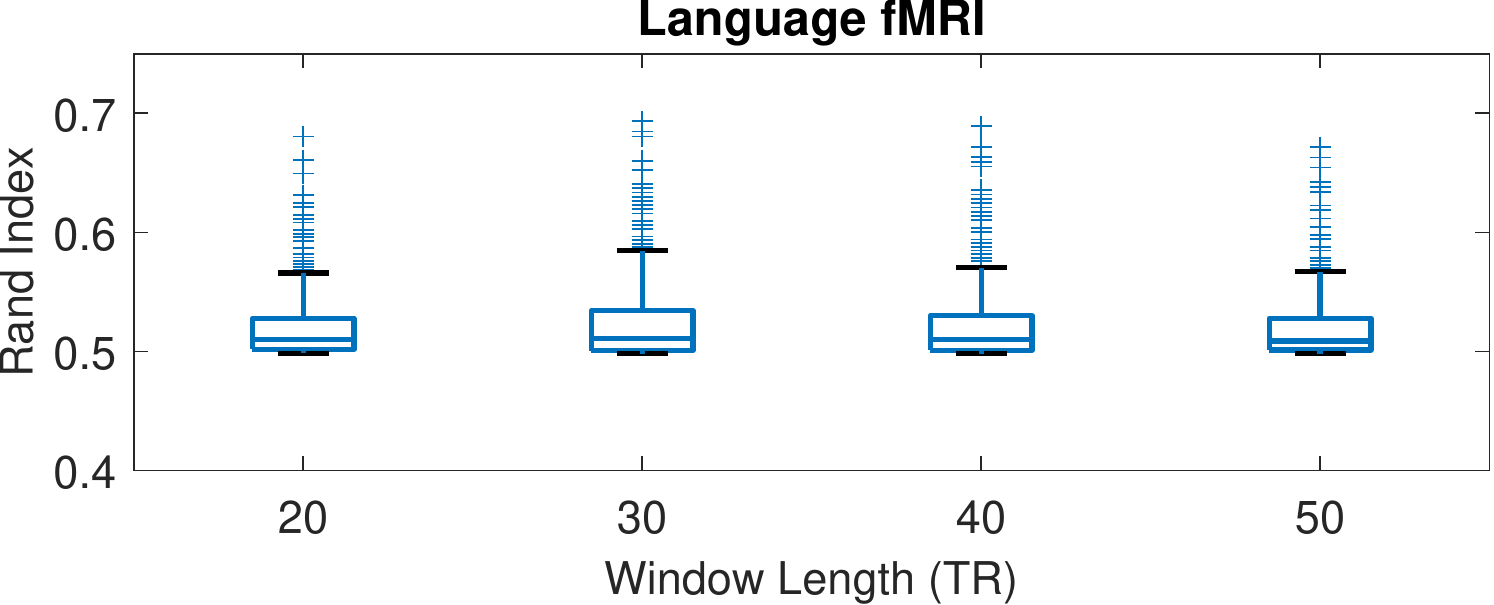}} \vspace{-0.2cm} \\
\subfigure[]{\includegraphics[width=0.6\textwidth,keepaspectratio]{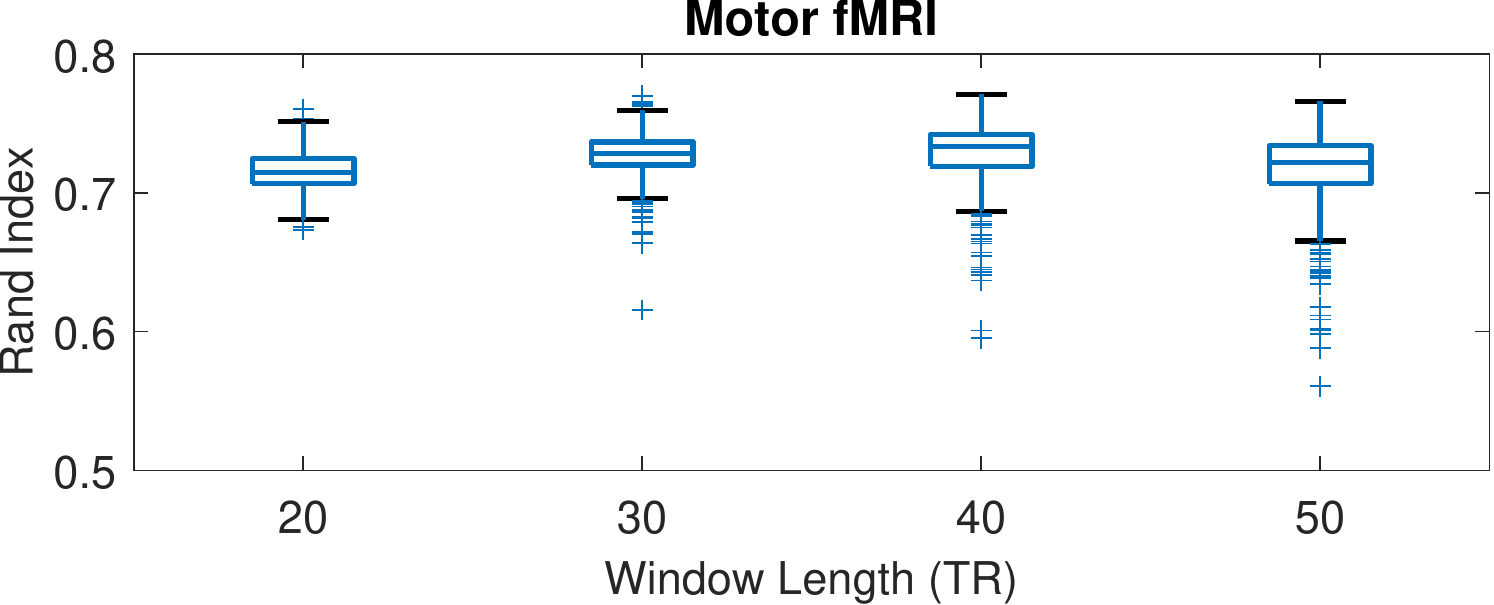}}
\vspace{-0.1in}
\caption{Effect of window length on the tracking of dynamic modular connectivity states in language and motor task fMRI data, as measured by RIs of the estimated and ground-truth state sequences over subjects.}
\label{Fig:Eff-wlen-ri}
\end{figure*}

\subsection{Effect of Window Length}

\begin{figure}[!t]
\centering
\includegraphics[width=0.5\linewidth,keepaspectratio]{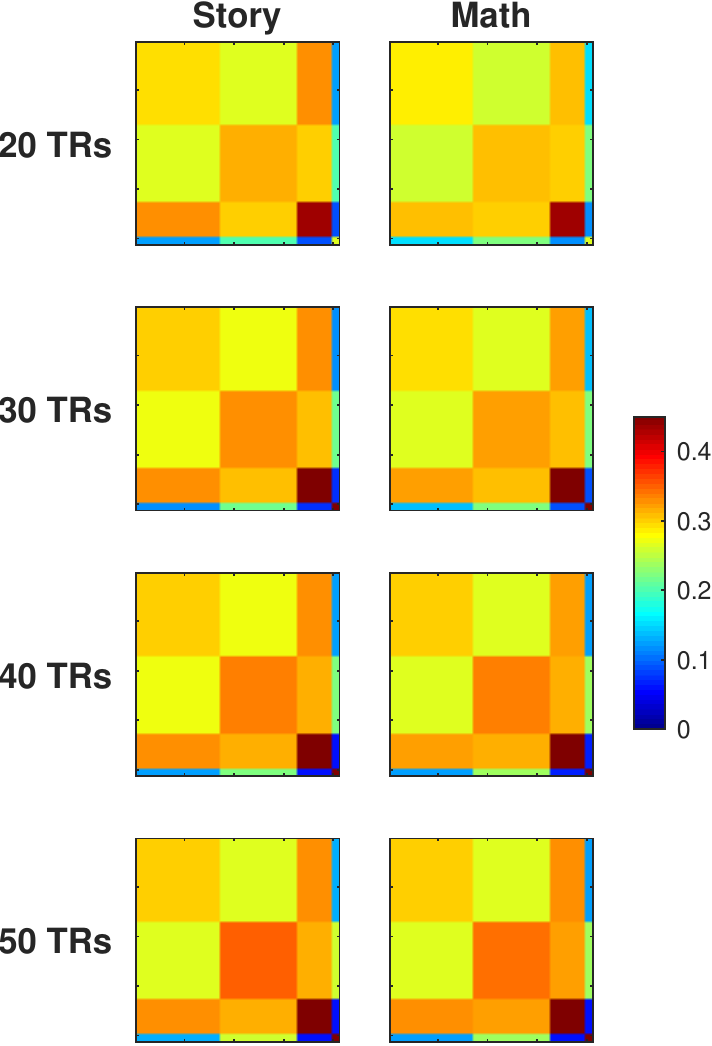} 
\vspace{-0.02in}
\caption{Effect of window length on the estimated state-based modular connection probability matrices for language tasks.} \label{Fig:Eff-wlen-mat-lang}
\end{figure}

\begin{figure*}[!t]
\centering
\includegraphics[width=0.95\linewidth,keepaspectratio]{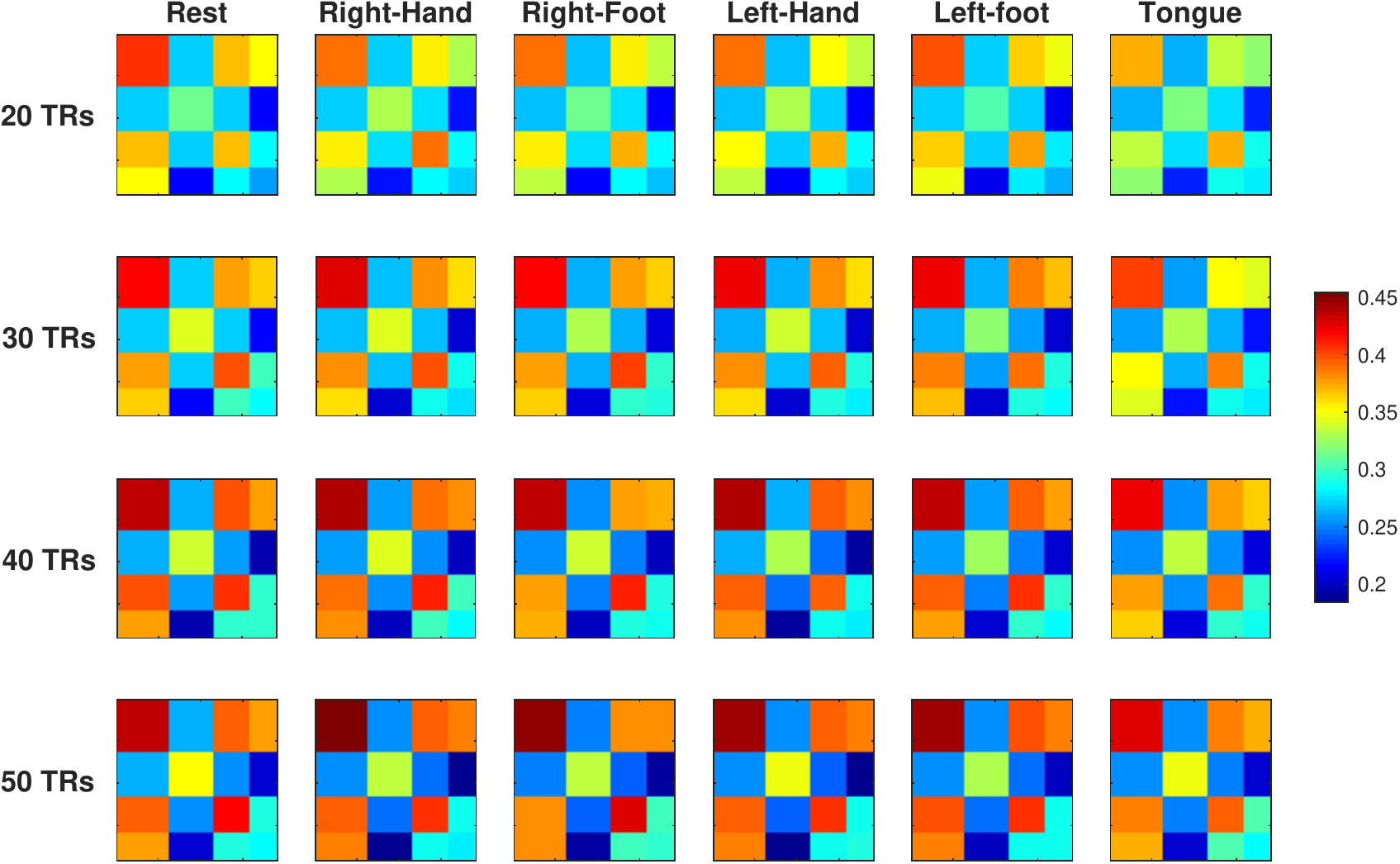} 
\vspace{-0.02in}
\caption{Effect of window length on the estimated state-based modular connection probability matrices for motor tasks.} \label{Fig:Eff-wlen-mat-motor}
\end{figure*}

Estimation of sliding-window correlations requires choosing an appropriate window length which is crucial: Too short a window produces noisy estimates of the correlation measures when derived from fewer data points, and introduces spurious fluctuations in the observed dynamic FC; Too long a window has low sensitivity in detecting highly-localized genuine FC variations \cite{Shakil2016,Preti2017}. To investigate the effect of window length on the dynamic modular connectivity estimation, we examined time windows of 20, 30, 40 and 50 TRs ($\approx$ 15, 22, 29 and 36 s for HCP data). Fig.~\ref{Fig:Eff-wlen-ri} plots the Rand indices (RIs) between the estimated and ground-truth (from experimental paradigm) state sequences as a function of window length for both tasks. As expected, there were observable differences in the dynamic state identification performance at different window lengths. We chose window length of 30 TRs which gives the highest RI of estimated state dynamics, indicating optimal sensitivity and specificity in detecting dynamic modular connectivity structure in these datasets. Nevertheless, the results suggest the estimates are fairly robust against the choice of window size, as indicated by the relatively stable RIs over increasing window length. This is also evident from the similar patterns of the estimated state-specific modular connection probability matrices (based on multi-subject community detection) across different window lengths, as shown in Fig~\ref{Fig:Eff-wlen-mat-lang} (language task) and Fig~\ref{Fig:Eff-wlen-mat-motor} (motor task). This is probably because the inter-modular connection probabilities are global summary measures of network connectivity (between communities) derived from local connectivity (between nodes), which are relatively stable over time and thus less sensitive to varied window length compared to node-wise dynamic connectivity.

\subsection{Determining Number of States}

Determining the number of states $S$ is a critical problem in dynamic FC state identification. For task fMRI as in our case, it is reasonable to assume that the number of dynamic community states corresponds to the number of tasks in the experiments. Therefore, in the main text, we reported results by setting $S=2$ and $S=6$ for the language and motor tasks, respectively. When $S$ is unknown \textit{a priori} (typically for the resting-state data), we suggest a data-driven procedure to find $S$ via the K-means clustering of the time-varying modular connectivity matrices $\widehat{\boldsymbol{\Theta}}^{r,t}$ as an initial step. The optimal number of clusters can be determined by cluster validity indices and serves as an estimate of $S$ \cite{Vergara2020}. This is a common approach to identifying unknown number of states for whole-brain dynamic FC in resting-state fMRI \cite{Allen2012,Shakil2016}. As in \cite{Samdin2017,Ting2018,Ombao2018}, we can also use the identified clusters to initialize the parameters of HMM states, where each cluster represents a state in the Markov chain. For illustration, Table.~\ref{Table:selK} shows the selected $S$ for our task fMRI using the silhouette and Davies-Bouldin indices. The silhouette method is widely used in dynamic FC studies, while the Davies-Bouldin index showed better performance in identifying $S$ in both simulation and real fMRI data among twenty four cluster indices tested recently in \cite{Vergara2020}. We evaluated a range of candidates $S \in \{1,2, \ldots, S_{\text{max}}\}$ with $S_{\text{max}}=5$ and $S_{\text{max}}=10$ for the language and motor tasks. Both indices yield reasonable estimates of $S$ which are in close agreement with the number of tasks especially for the motor task. A slightly better approximation was obtained with the subject-specific community partition compared to the multi-subject. This was likely due to model fitting to individual subjects which allows for varied dimensions of $\widehat{\boldsymbol{\Theta}}^{r,t}$ used in the clustering.

\begin{table}[!t]
\renewcommand{\arraystretch}{1.3}
\caption{Means and standard deviations (in parenthesis) of selected number of states $\hat{S}$ in time-evolving modular connectivity over individual subjects via cluster analysis.}
\vspace{-0.1 cm}
\label{Table:selK}
\centering
\begin{tabular}{p{3cm}p{4cm}cc}
\hline \hline
\multirow{2}{*}{Task fMRI} & \multirow{2}{*}{Community Partition} & \multicolumn{2}{c}{Cluster Validity Indices} \\
\cline{3-4}
& & Silhouette & Davies-Bouldin \\
\hline
\multirow{2}{*}{Language} & Subject-specific & 3.19 (1.15) & 3.23 (1.23) \\
& Multi-subject & 3.49 (1.20) & 3.93 (1.20) \\
\multirow{2}{*}{Motor} & Subject-specific & 6.09 (2.90) & 6.43 (3.14) \\
& Multi-subject & 6.20 (2.96) & 6.66 (3.17) \\
\hline \hline
\end{tabular}
\vspace{0.2in}
\end{table}

\section{Convergence Analysis}

\begin{figure}[!t]
\centering
\includegraphics[width=0.5\linewidth,keepaspectratio]{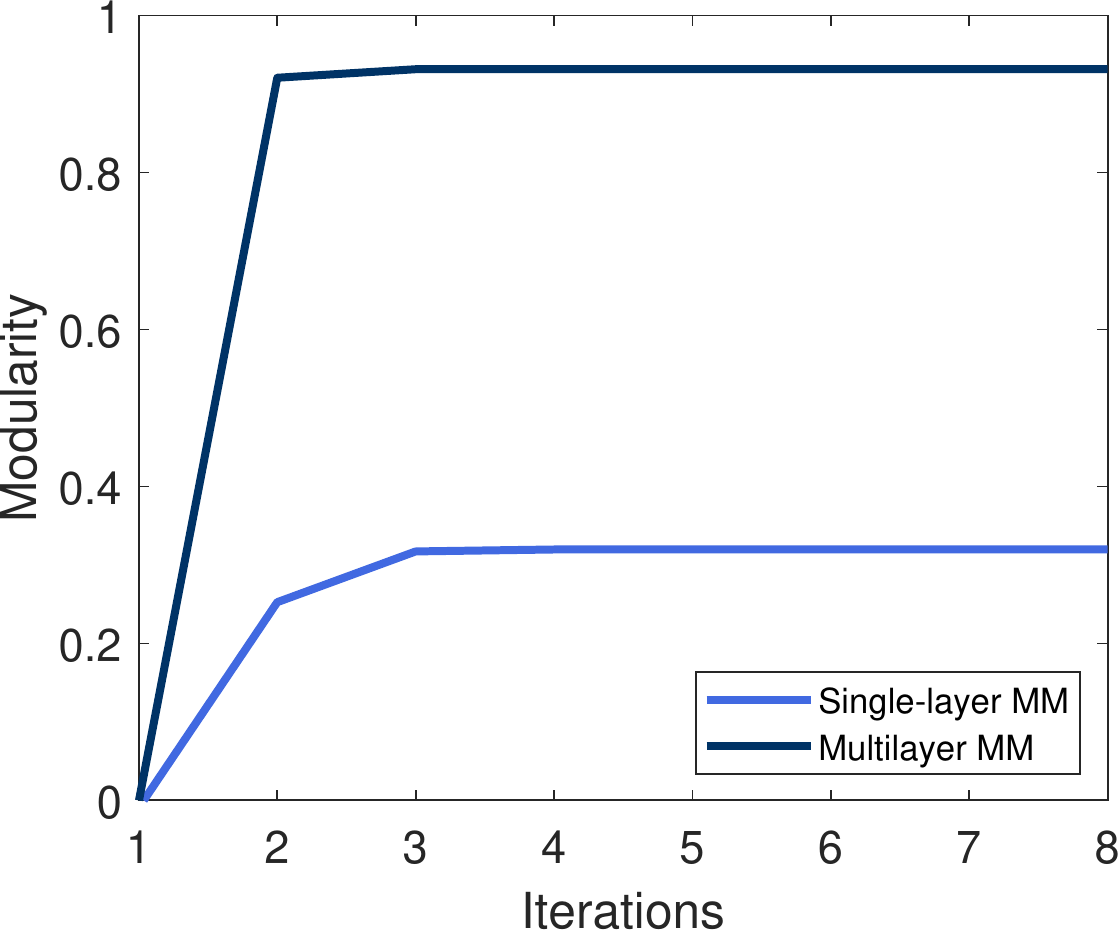} 
	\caption{Convergence behavior of generalized Louvain algorithm for single-subject and multi-subject community detection on one realization of simulated networks with $N=120$, $K=12$ and $R=100$. The plot for single-subject analysis is average over $R=100$ subjects.} \label{converg}
\end{figure}

We investigate the convergence behavior of the generalized Louvain algorithm for community detection in multi-subject networks via the modularity maximization. Fig.~\ref{converg} plots the modularity index $Q$ computed at each iteration of the algorithm on simulated networks. The modularity increases monotonically as iteration increases, converging rapidly at first few iterations for both the single-layer and multilayer community detection. Nevertheless, the algorithm for the multilayer modularity maximization shows slightly faster convergence to a higher modularity value, due to pooling of shared information from networks across subjects in the optimization to detect a common community structure.

\section{Computational Time}

In Table.~\ref{Table:Comptime}, we compare the computational time of different algorithms in the community detection and dynamic modular connectivity state identification for the language and motor tasks (using MATLAB R2019a on MacBookPro laptop with 3.1 GHz Intel Core i5 CPU and 8 GB memory). The computational time required to estimate community partition in functional networks over all subjects was substantially lower in the group-level analysis compared to subject-specific analysis. This was probably due to the faster convergence of the multilayer modularity maximization algorithm as shown in Fig.~\ref{converg}. As expected, the use of HMM in identifying dynamic connectivity states was computationally more intensive than the K-means clustering, which however gave more accurate tracking of dynamic states for both tasks (Fig. 9, main text) and provided additional information on the temporal organization of the inferred brain states specified by the state transition probability matrices (Fig. 7, main text). The computational cost for state identification increased as the number of states and the dimension of the modular connectivity matrices increase. Hence, the computational cost related to modeling motor tasks (with more states and a larger number of detected communities) was substantially greater compared to that of the language tasks.

\begin{table}[!th]
\vspace{0.2in}
\renewcommand{\arraystretch}{1.3}
\caption{Computational time (in seconds) for community detection (over all subjects) and dynamic modular connectivity state identification (per subject) for the language and motor tasks using different methods.}
\vspace{-0.1 cm}
\label{Table:Comptime}
\centering
\begin{tabular}{p{3cm}cccc}
\hline \hline
\multirow{2}{*}{Task fMRI} & \multicolumn{2}{c}{Community Detection} & \multicolumn{2}{c}{State Identification}\\
& Subject-Specific & Multi-Subject & K-means & HMM \\
\hline
Language & 13.06 & 0.22 & 0.0034 & 5.94  \\
Motor & 13.98 & 0.24 & 0.0062 & 45.99 \\
\hline \hline
\end{tabular}
\vspace{0.2in}
\end{table}

\bibliographystyle{IEEEbib}
\bibliography{SBM-Ref-IEEE}